\documentclass[runningheads]{llncs}
\usepackage{graphicx}
\usepackage{comment}

\usepackage{amsmath,amssymb,amsfonts,amsthm,bm} 
\usepackage{color}

\usepackage{graphicx}
\usepackage{subcaption}
\usepackage{float}
\usepackage{caption}	
\usepackage{lscape}     

\usepackage[lined,ruled,linesnumbered]{algorithm2e}

\usepackage{booktabs}                   
\usepackage{multirow}

\usepackage{paralist}
\usepackage{enumitem}

\usepackage{bm}                          
\usepackage{epsfig}                      
\usepackage{graphicx}                  
\usepackage{mathtools}

\usepackage{color}

\usepackage{comment}

\usepackage{url}  
\usepackage[pagebackref=false,breaklinks=true,colorlinks=true,filecolor=blue,urlcolor=blue,linkcolor=red,bookmarks=false]{hyperref}
\usepackage[nocompress]{cite}

\usepackage{listings}

\usepackage{xspace}
\usepackage[table]{xcolor}
\usepackage{setspace}

\usepackage{nicefrac}
\usepackage{microtype}
\usepackage[utf8]{inputenc} 
\usepackage[T1]{fontenc}    

\def\etal{et~al.}			  
\def\eg{e.g.,~}               
\def\ie{i.e.,~}               

\newcommand{\para}[1]{\vspace{1mm}\noindent\textbf{#1}}

\newlength\paramargin
\newlength\figmargin
\newlength\secmargin
\newlength\figcapmargin
\newlength\tabmargin

\newcommand {\first}[1]{{\color{red}\textbf{#1}}}
\newcommand {\second}[1]{{\color{blue}\underline{#1}}}

\newcommand{\mpage}[2]
{
\begin{minipage}{#1\linewidth}\centering
#2
\end{minipage}
}

\newcommand{\heading}[1]
{
\vspace{1mm}\noindent\textbf{#1}
}

\newcommand{\secref}[1]{Section~\ref{sec:#1}}
\newcommand{\figref}[1]{Figure~\ref{fig:#1}} 
\newcommand{\tabref}[1]{Table~\ref{tab:#1}}

\long\def\ignorethis#1{}

\newcommand {\yl}[1]{{\color{red}\textbf{YL: }#1}\normalfont}

\newcommand{\tb}[1]{\textbf{#1}}



\def\xi{\mathbf{x}_i}

\graphicspath{{figure}, {example}}

\usepackage[width=122mm,left=12mm,paperwidth=146mm,height=193mm,top=12mm,paperheight=217mm]{geometry}

\begin{document}
\pagestyle{headings}
\mainmatter
\def\ECCVSubNumber{1714}  

\title{DRG: Dual Relation Graph for\\ Human-Object Interaction Detection}
\titlerunning{DRG: Dual Relation Graph for Human-Object Interaction Detection}

\author{
Chen Gao \and
Jiarui Xu \and
Yuliang Zou \and
Jia-Bin Huang
}
\authorrunning{C. Gao et al.}
\institute{Virginia Tech \\
\email{\{chengao,jiaruixu,ylzou,jbhuang\}@vt.edu}}


\maketitle
\newlength\ffca
\setlength\ffca{4cm} 
\newlength\ffcb
\setlength\ffcb{-0.0mm}

\begin{center}
\parbox[t]{\ffca}{
\centering%
\includegraphics[width=\ffca]{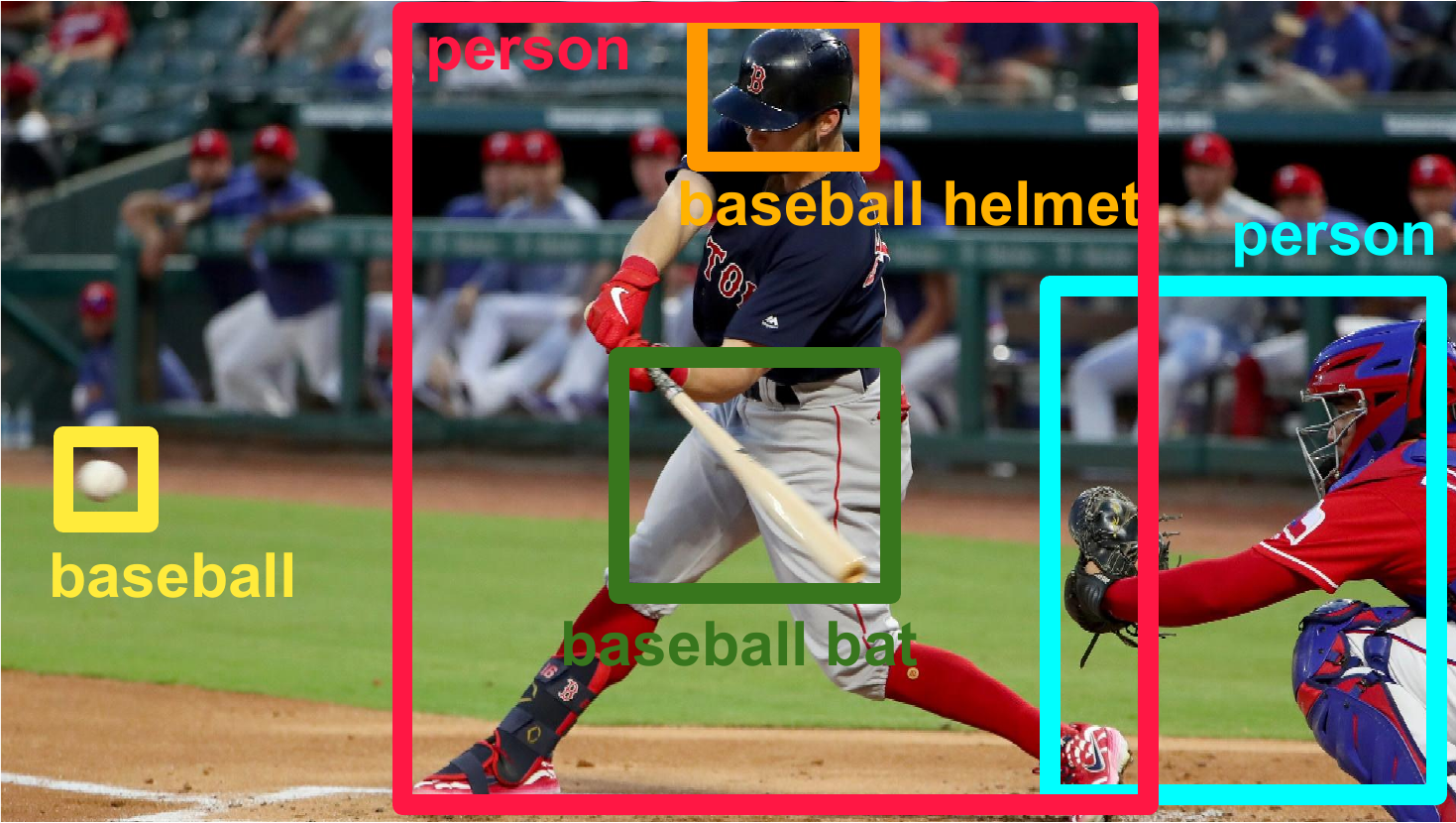}\\%
 (a) Object detection}%
\hfill%
\parbox[t]{\ffca}{
\centering%
\includegraphics[width=\ffca]{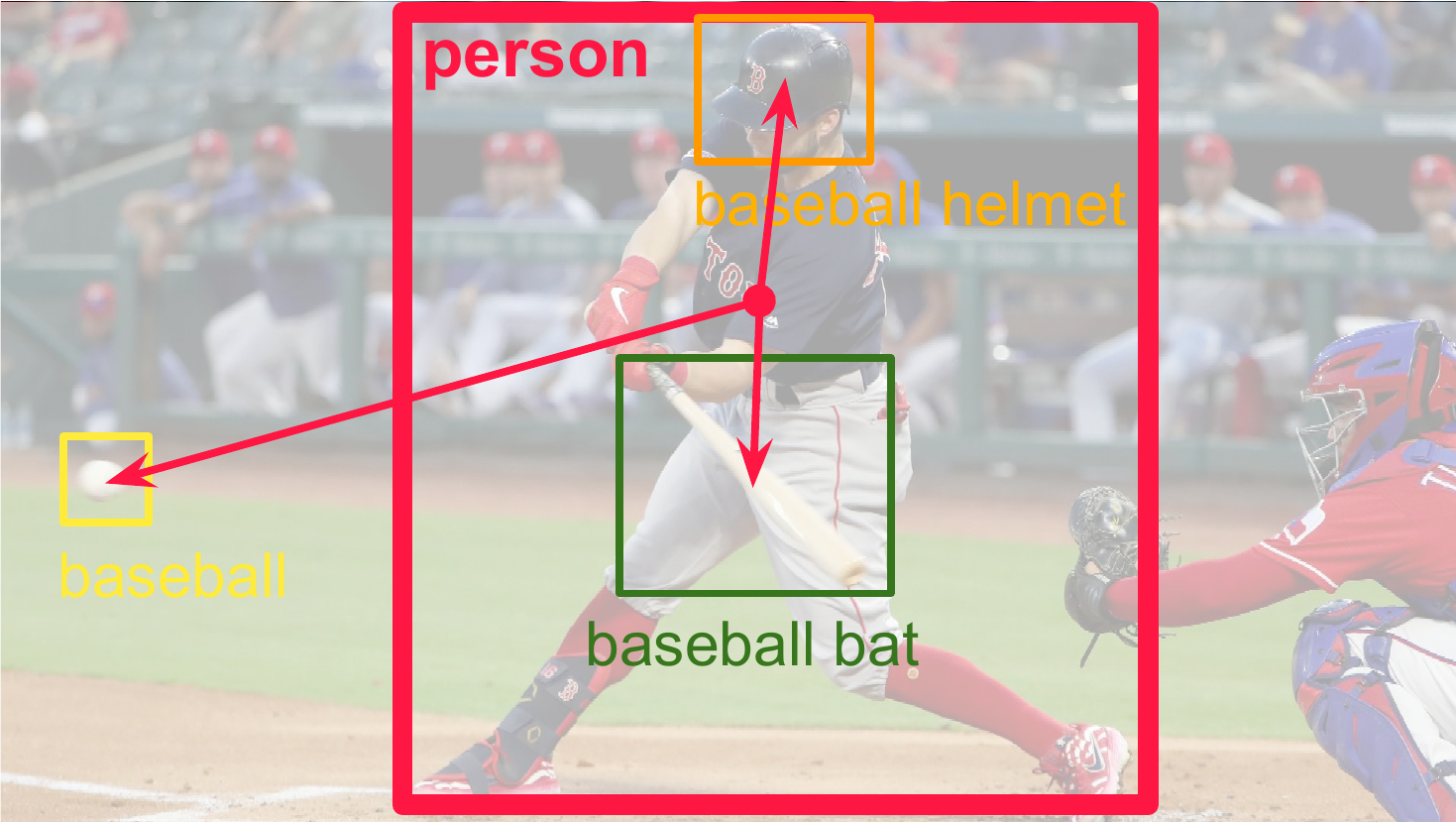}\\%
 (b) Human-centric}%
\hfill%
\parbox[t]{\ffca}{
\centering%
\includegraphics[width=\ffca]{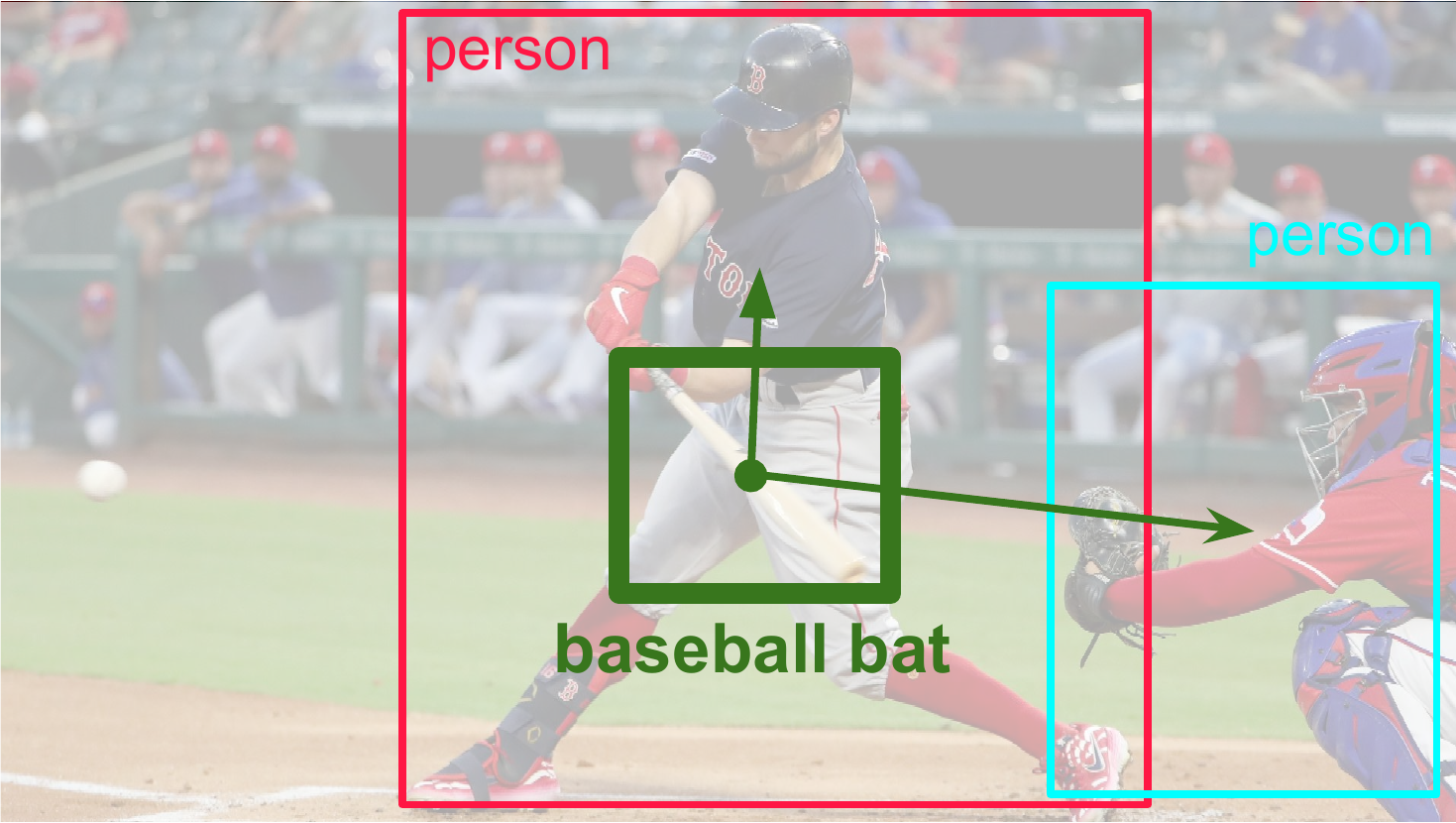}\\%
(c) Object-centric}%

  \captionof{figure}{
    \textbf{Human-object interaction (HOI) detection using dual relation graph.}
    Predicting each HOI in isolation is ambiguous due to the lack of context. 
    In this work, we propose to leverage a dual relation graph. 
    %
    For each human node \textit{h}, we obtain a \emph{human-centric} subgraph where all object nodes are connected to \textit{h}.
    Similarly, we can obtain an \emph{object-centric} subgraph for each object node \textit{o}.
    %
    The human subgraph helps to adjust single HOI's prediction based on the same person's other HOIs. For example, knowing a 
    \textbf{\textcolor[HTML]{ff1744}{person}} 
    is wearing a 
    \textbf{\textcolor[HTML]{ff9900}{baseball helmet}}
    and hitting a 
    \textbf{\textcolor[HTML]{dfcf44}{baseball}} 
    suggests that the 
    \textbf{\textcolor[HTML]{ff1744}{person}} 
    may be holding a 
    \textbf{\textcolor[HTML]{38761d}{baseball bat}}.
    Similarly, the object subgraph helps to refine the HOI's prediction based on other HOIs associated with the same object. 
    For example, knowing a 
    \textbf{\textcolor[HTML]{38761d}{baseball bat}} 
    is held by a 
    \textbf{\textcolor[HTML]{ff1744}{person}} 
    lowers the chance that it is held by another 
    \textbf{\textcolor[HTML]{00ffff}{person}}.
    Our method exploits such cues for improving HOI detection.
  }
  \label{fig:teaser}
\end{center}


\begin{abstract}
%
We tackle the challenging problem of human-object interaction (HOI) detection.
Existing methods either recognize the interaction of each human-object pair in isolation or perform joint inference based on complex appearance-based features.
In this paper, we leverage an abstract spatial-semantic representation to describe each human-object pair and aggregate the contextual information of the scene via a dual relation graph (one \emph{human-centric} and one \emph{object-centric}). 
Our proposed dual relation graph effectively captures discriminative cues from the scene to resolve ambiguity from local predictions.
Our model is conceptually simple and leads to favorable results compared to the state-of-the-art HOI detection algorithms on two large-scale benchmark datasets.
\end{abstract}

%
\section{Introduction}
\label{sec:intro}
Detecting individual persons and objects in isolation often does not provide sufficient information for understanding complex human activities.
Moving beyond detecting/recognizing individual objects, we aim to detect persons, objects, and recognize their interaction relationships (if any) in the scene.
This task, known as human-object interaction (HOI) detection, can produce rich semantic information with visual grounding.

State-of-the-art HOI detection methods often use appearance features from the detected human/object instances as well as their relative spatial layout for predicting the interaction relationships~\cite{Chao-WACV-HOI,Gao-BMVC-iCAN,Gkioxari-CVPR-InteractNet,Gupta-SemanticRoleLabeling,Alex-No-Frills,Kolesnikov-BAR,Li-CVPR-Interactiveness, wang2019deep,zhou2019relation,Bo-PMFNet,peyre2018detecting,bansal2019detecting,liao2020ppdm}. 
These methods, however, often predict the interaction relationship between each human-object pair \emph{in isolation}, thereby ignoring the contextual information in the scene.
In light of this, several methods have been proposed to capture the contextual cues through iterative message passing~\cite{Qi-ECCV-GraphParsing,Xu-CVPR-MessagePassing} or attentional graph convolutional networks~\cite{Yang-ECCV-Graph-RCNN}.
However, existing approaches rely on complex appearance-based features to encode the human-object relation (e.g., deep features extracted from a union of two boxes) and do not exploit the informative spatial cues.
In addition, the contexts are aggregated via a \emph{densely connected} graph (where the nodes represent all the detected objects).

%
In this paper, we first propose to use spatial-semantic representation to describe each human-object pair.
Specifically, our spatial-semantic representation encodes (1) the relative spatial layout between a person and an object and (2) the semantic word embedding of the object category.
Using spatial-semantic representation for HOI prediction has two main advantages:
First, it is invariant to complex appearance variations.
Second, it enables knowledge transfer among object classes and helps with rare interaction during training and inference.

While such representations are informative, predicting HOI in isolation fails to leverage the contextual cues.
In the example of \figref{teaser}, a model might struggle to recognize that the person (in the red box) is hitting the baseball, by using only the spatial-semantic features from this particular human-object pair.
Such ambiguity, however, may be alleviated if given the relation among different HOIs \emph{from the same person}, e.g., this person is wearing a baseball helmet and holding a baseball bat.
Similarly, we can exploit the relations among different HOIs \emph{from the same object}.
For example, a model may recognize both persons are holding the same baseball bat when making prediction independently.
Knowing that the person (red box) is more likely to hold the baseball bat reduces the probability of another person (blue box) holding the same baseball bat. 
Inspired by these observations, we construct a \emph{human-centric} and an \emph{object-centric} HOI subgraph and apply attentional graph convolution to encode and aggregate the contextual information.
We refer to our method as Dual Relation Graph (DRG).

\para{Our contributions.} 
\begin{compactitem}
\item We propose Dual Relation Graph, an effective method to capture and aggregate contextual cues for improving HOI predictions.

\item We demonstrate that using the proposed spatial-semantic representation alone (without using appearance features) can achieve competitive performance compared to the state-of-the-art.

\item We conduct extensive ablation study of our model, identifying contributions from individual components and exploring different model design choices.

\item We achieve competitive results compared with the state-of-the-art on the VCOCO and HICO-DET datasets.
\end{compactitem}

\ignorethis{
%
%
%
This abstract representation will show its power on zero-shot learning. Furthermore, we believe naturally it is a perfect fit for the zero-shot task.

Recent years have witnessed rapid progress in visual recognition tasks, including object detection~\cite{Dai-NIPS-RFCN,Ross-CVPR-FastRCNN,Lin-CVPR-Pyramid,Ren-NIPS-FasterRCNN}, semantic and instance segmentation~\cite{Chen-TPAMI-DeepLab,Ross-CVPR-Hierarchies,He-ICCV-MaskRCNN,Long-CVPR-FCN}, and action recognition~\cite{Cheron-ICCV-PCNN,Girdhar-NIPS-AttentionalPooling,Maji-CVPR-Action}.
%
These tasks, however, focus on characterizing \emph{individual} object instance, while neglecting the rich semantic information among these instances. 
Recently, the scene graph, a visually-grounded graphical structure of an image, is used to model the visual relationship among individual object instances~\cite{Hu-CVPR-Relation,Li-ICCV-SceneGraph,Newell-NIPS-Pixels,Xu-CVPR-MessagePassing,Yang-ECCV-Graph-RCNN,Yang-ECCV-Shuffle,Zellers-CVPR-Neural}, which leads to a deeper understanding of the scene.

%
In this paper, we study a particularly important class of visual relationship detection, which detects how each person interacts with the surrounding objects. This task, known as Human-Object Interaction (HOI) detection~\cite{Chao-WACV-HOI,Gupta-SemanticRoleLabeling}, aims to localize a person, an object, as well as identify the interaction between the person and the object.
%
Given an image, we want to detect all the $\langle$ $\texttt{human, verb, object}$ $\rangle$ triplets.

Understanding Human-Object Interactions (HOIs) is an essential step towards a deeper understanding of the scene. Since most of the images are \emph{human-centric}, studying the HOI detection problem provides essential cues for high-level scene understanding tasks, such as pose estimation~\cite{Cao-CVPR-OpenPese,Yao-BMVC-Pose}, image captioning~\cite{Li-CVPR-Scene,Lu-CVPR-Baby}, and image retrieval~\cite{Johnson-CVPR-Retrieval}.


With the rapid development of deep convolutional neural networks (DCNs), several methods~\cite{Chao-WACV-HOI,Gao-BMVC-iCAN,Gkioxari-CVPR-InteractNet,Gupta-SemanticRoleLabeling,Qi-ECCV-GraphParsing} have been proposed to tackle the problem.
These approaches, however, do not effectively encode the object layout information, which we argue is an informative cue for HOI detection.
Gkioxari~\etal~\cite{Gkioxari-CVPR-InteractNet} incorporates this geometric cue by regressing object location from human bounding box with a Mixture Density Network. Chao~\etal~\cite{Chao-WACV-HOI} encodes the relative location information using a two-channel binary image representation.
\yl{Not very sure about the claim.}
Both methods ignore the object category information when encoding the location information, which leads to a sub-optimal solution.
We argue that incorporating object category information could capture the location relation more effectively.
Thus, we explore a spatial-semantic feature representation by combining the spatial relation between a person and an object, and the object category information (encoded with pre-trained word embedding).
In~\figref{teaser}, we show an example of detecting human-object interaction from the spatial layout between a person and an object, as well as the corresponding object categories. 
However, this local information is not discriminative enough and will cause ambiguities.
Though the two persons highlighted in the white boxes share similar human-object spatial layout (\emph{center}), they are performing different actions (\emph{left}).
Thus, we leverage the \emph{global object layout} in the image (\emph{right}), which provides discriminative cues for inferring the interaction. Our intuition is that the global object layout serves as a complement to an individual HOI.
In the first example, the spatial layout between the person and the bench, the bag, and another person nearby together suggests that the person may be resting. 
In the second example, the spatial layout of the person and tennis ball strongly indicator of hitting action.
In this paper, we propose the Spatial-Semantic Aware Relation (SSAR) module to aggregate features from a set of HOI instances, \eg all the interaction between the same person and other surrounding objects.
Inspired by~\cite{Hu-CVPR-Relation}, we adopt an attention mechanism to model the dependencies among HOIs. We measure the similarity between different HOIs in an end-to-end learnable embedding space and use it to get a weighted sum of the input features. 
Note that we can aggregate HOIs features iteratively by stacking more numbers of SSAR module.
We argue that using this non-appearance feature can achieve a reasonably good performance.

%

\paragraph{Our contributions.}

We make the following contributions.
\begin{itemize}
\item We propose the Spatial-Semantic Aware Relation (SSAR) module to refine interaction prediction iteratively 
\item We construct a strong baseline for HOI detection by using the spatial-semantic representation only.
\item With the full model, we establish new state-of-the-art performances on two HOI detection benchmarks
\item We conduct detailed ablation experiments to identify the relative contributions of each individual component.
\item We will release all the source code and pre-trained models to facilitate future research.
\end{itemize}
}
\section{Related Work}
\label{sec:related}
\newlength\ffma
\setlength\ffma{3cm} 
\newlength\ffmb
\setlength\ffmb{-0.0mm}

\begin{figure}[t]
\centering%

\parbox[t]{\ffma}{
\centering%
\includegraphics[width=\ffma]{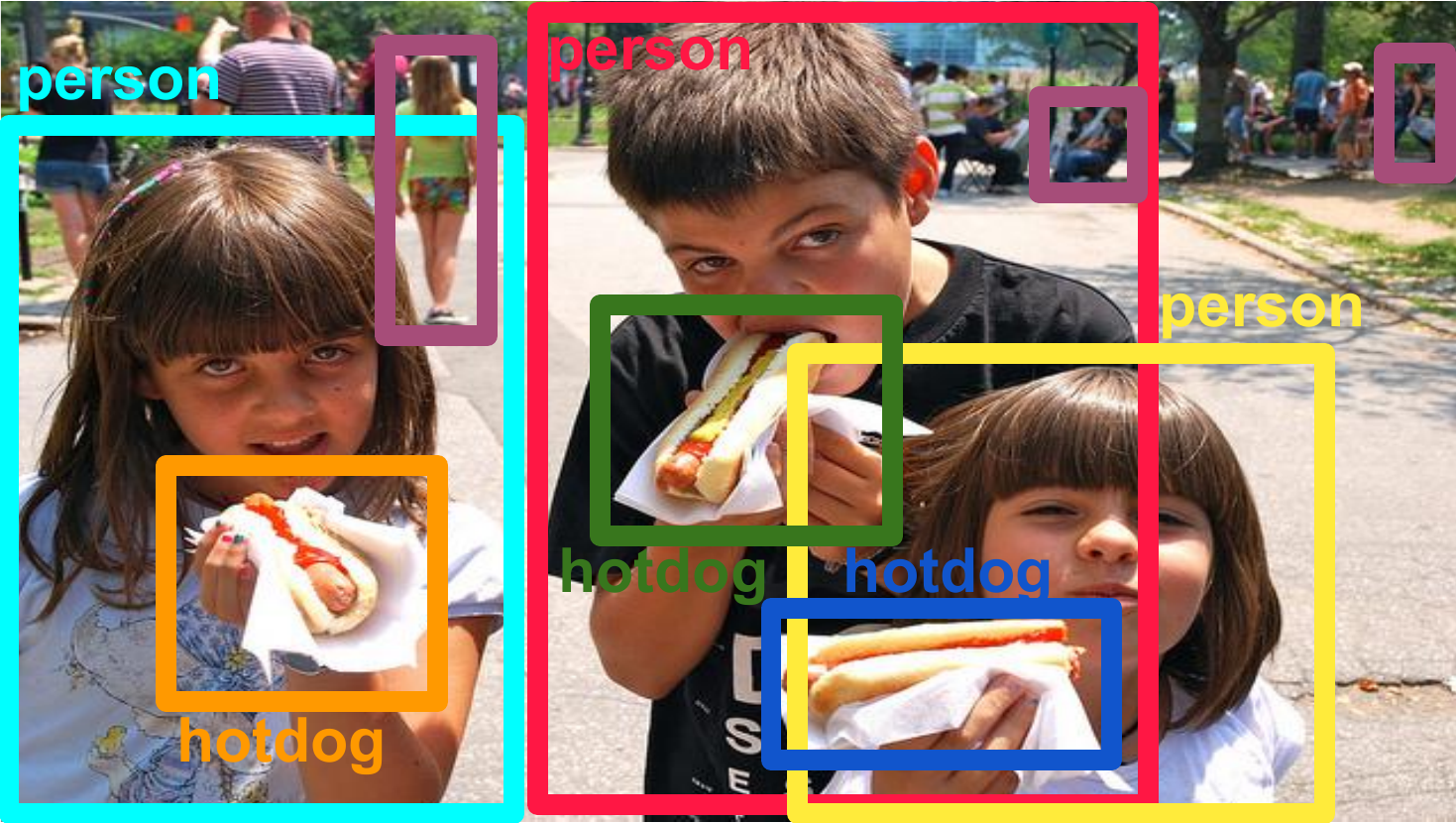}\\%
 \tiny (a) Object detection}%
\hfill%
\parbox[t]{\ffma}{
\centering%
\includegraphics[width=\ffma]{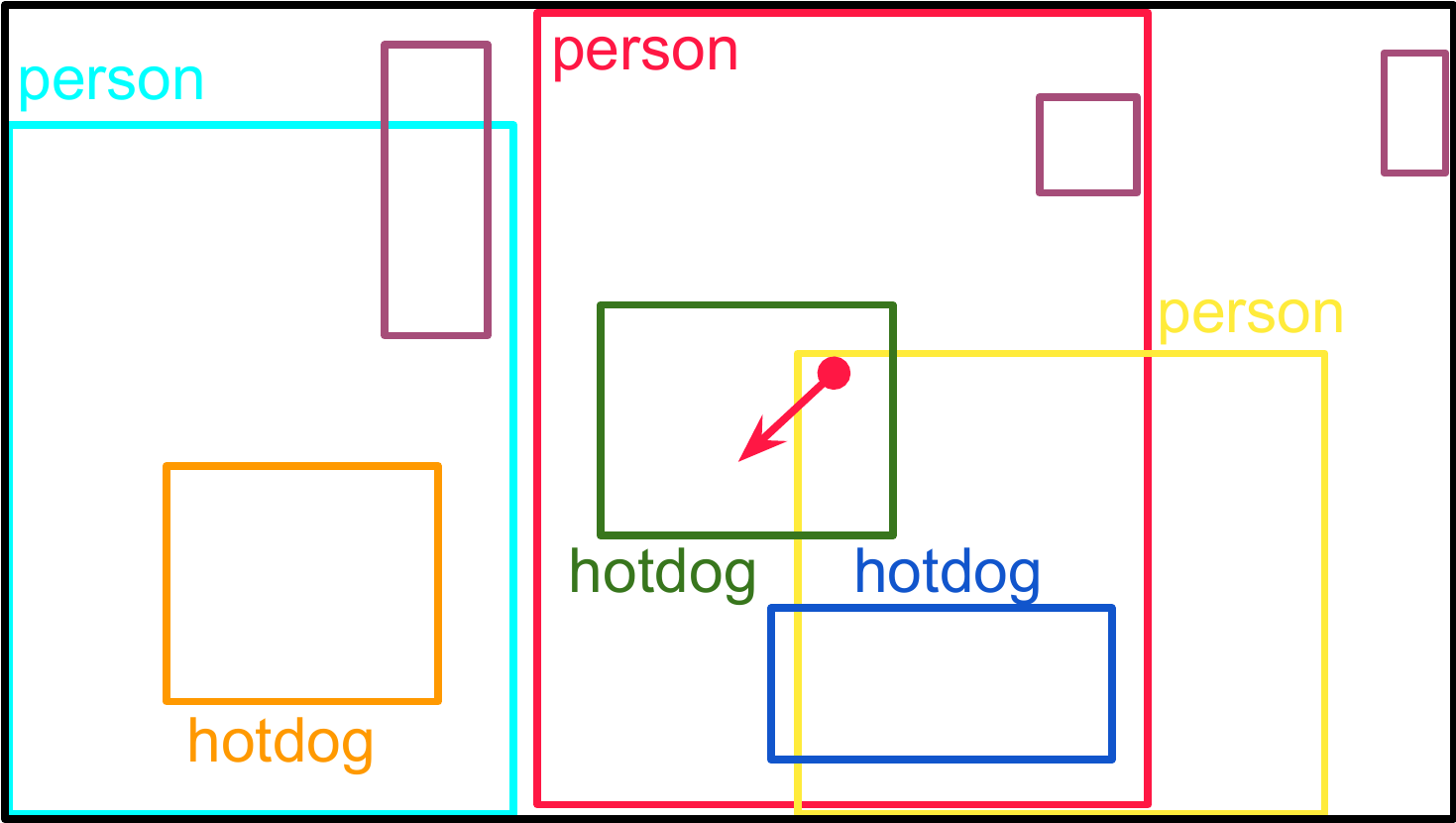}\\%
 \tiny (b) Independent}%
\hfill%
\parbox[t]{\ffma}{
\centering%
\includegraphics[width=\ffma]{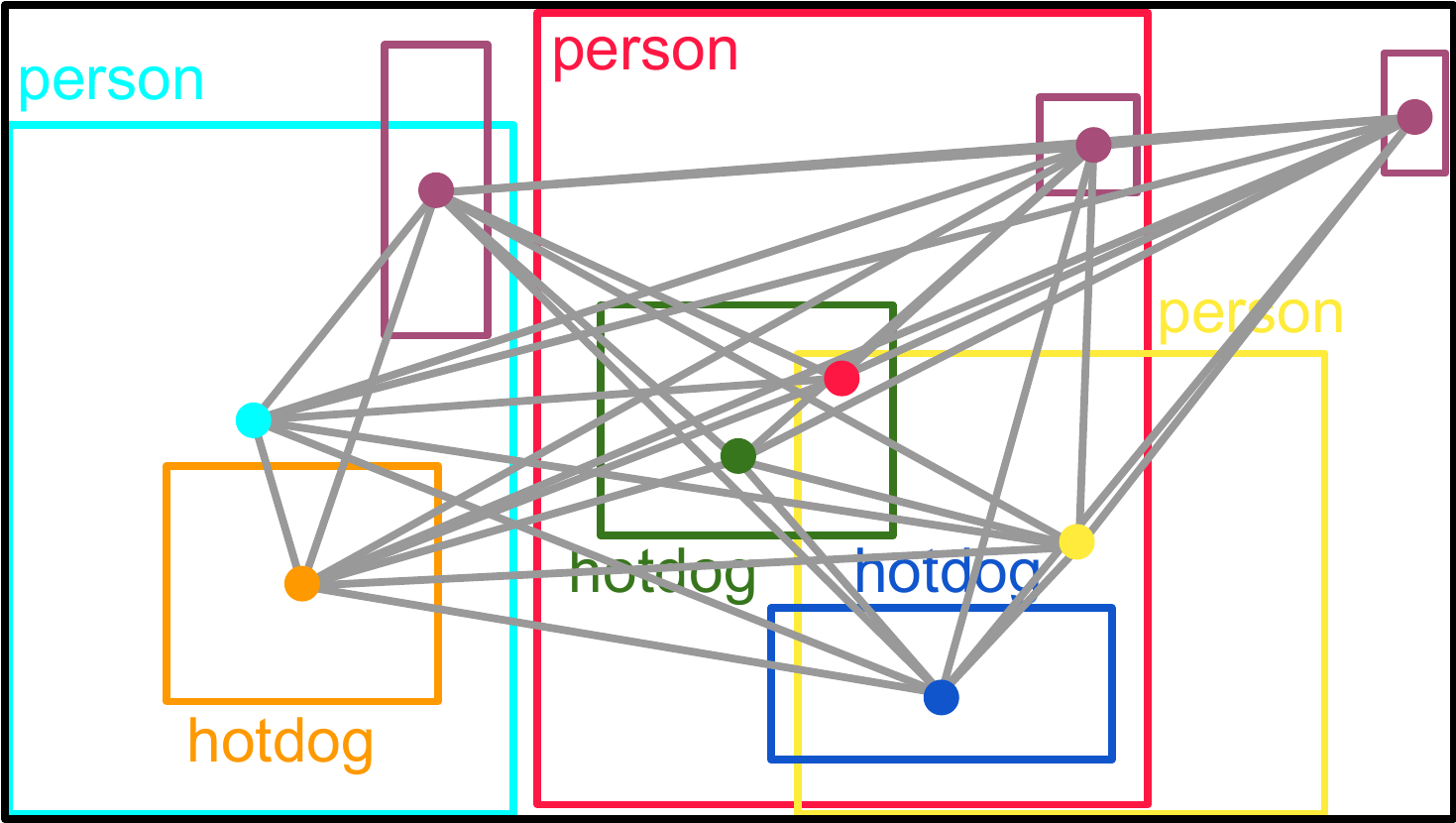}\\%
 \tiny (c) Densely connected}%
\hfill%
\parbox[t]{\ffma}{
\centering%
\includegraphics[width=\ffma]{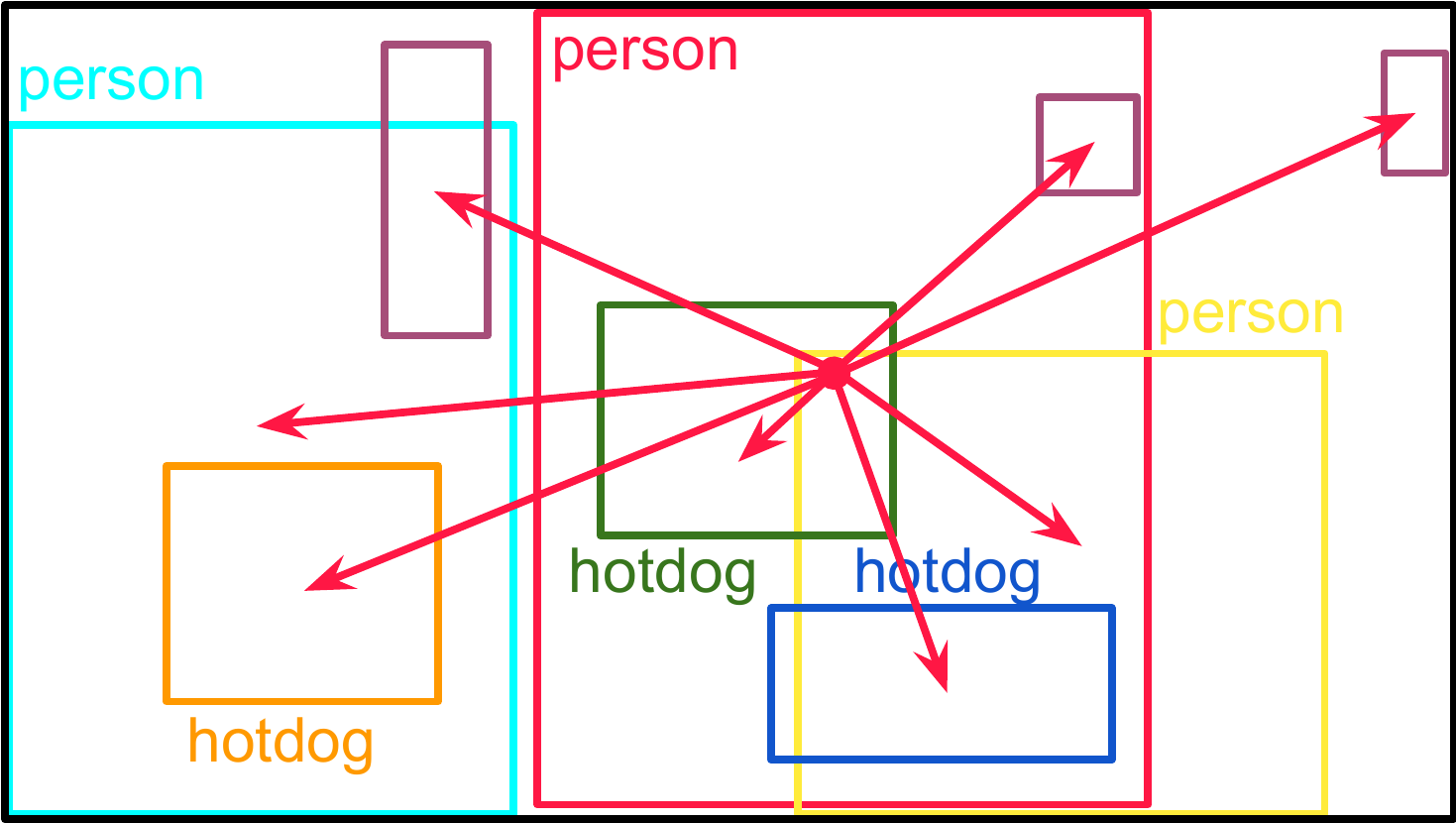}\\%
 \tiny (d) Ours (human-centric)}%

  \captionof{figure}{
\textbf{Leveraging contextual information.} Given object detections in the scene (a), existing HOI detection algorithms only perform \emph{independent} prediction for each Human-object pair (b), ignoring the rich contextual cues. Recent methods in visual relationship detection (or scene graph generation) perform \emph{joint inference} on a densely connected graph (c). While being general, the large number of relations among the dense connections makes the learning and inference on such a graph challenging. 
In contrast, our work leverages the human/object-centric graph to focus only on relevant contexts for improved HOI detection (d). 
}
  \label{fig:motivation}
\end{figure}

\para{Human-object interaction detection.} 
The task of human-object interaction detection aims to localize persons, object instances, as well as recognize the interactions (if any) between each pair of a person and an object.
State-of-the-art HOI detection algorithms generally rely on two types of visual cues: 
(1) appearance features of the detected persons and objects (e.g., using the ROI pooling features extracted from a ConvNet) and 
(2) the spatial relationship between each human-object pair (e.g., using the bounding box transformation between the agent and the object~\cite{Gkioxari-CVPR-InteractNet,Gupta-SemanticRoleLabeling,Alex-No-Frills}, a two-channel interaction pattern~\cite{Chao-WACV-HOI,Gao-BMVC-iCAN}, or modeling the mutual contexts of human pose and object~\cite{Alex-No-Frills,Li-CVPR-Interactiveness,yao2010modeling}).
Recent advances focus on incorporating \emph{contexts} to resolve potential ambiguity in interaction prediction based on independent human-object pairs, including pairwise body-parts~\cite{Bo-PMFNet,fang2018pairwise} or object-parts~\cite{zhou2019relation}, instance-centric attention \cite{Gao-BMVC-iCAN,wang2019deep}, or message passing on a graph~\cite{Qi-ECCV-GraphParsing}.
Our work shares similar spirits with these recent efforts as we also aim to capture contextual cues.
The key difference lies in that the above approaches learn to aggregate contextual information from the other objects, body parts, or the scene background, while our method exploits  \emph{relations among different HOIs} to refine the predictions.

Inspired by the design of two-stage object detectors~\cite{Ren-NIPS-FasterRCNN}, recent works also show that filtering out candidate pairs with no relations using a relation proposal network~\cite{Yang-ECCV-Graph-RCNN} or an interactiveness network~\cite{Li-CVPR-Interactiveness} improves the performance.
Our method does not train an additional network for pruning unlikely relations.
We believe that incorporating such a strategy may lead to further improvement.

Recent advances in HOI detection focus on tackling the long-tailed distributions of HOI classes. 
Examples include transferring knowledge from seen categories to unseen ones by an analogy transformation~\cite{Peyre-ICCV-Weakly}, performing data augmentation of semantically similar objects~\cite{bansal2019detecting}, or leveraging external knowledge graph~\cite{Kato_2018_ECCV}.
While we do not explicitly address rare HOI classes, our method shows a small performance gap between rare and non-rare classes. 
We attribute this to the use of our abstract spatial-semantic representation.

%
%


\para{Visual relationship detection.} 
Many recent efforts have been devoted to detecting visual relationships~\cite{Bilen-CVPR-Weakly,Dai-CVPR-Relationship,Hu-CVPR-Referential,Li-CVPR-VIP,Peyre-ICCV-Weakly,Zhang-ICCV-PPR,Zhuang-ICCV-ContextAware}.
Unlike object detection, the number of relationship classes can be prohibitively large due to the compositionality of object pairs, predicates, and limited data samples.
To overcome this issue, some forms of language prior have been applied~\cite{Lu-ECCV-Prior,Plummer-ICCV-Phrase}.
Our focus in this work is on one particular class of relationship: human-centric interactions.
Compared with other object classes, the possible interactions (the predicate) between a person and objects are significantly more fine-grained.

\para{Scene graph.} 
A scene graph is a graphical structure representation of an image where objects are represented as nodes, and the relationships between objects are represented as edges~\cite{Newell-NIPS-Pixels,Xu-CVPR-MessagePassing,Yang-ECCV-Graph-RCNN,Yang-ECCV-Shuffle,Zellers-CVPR-Neural}. 
As the scene graph captures richer information beyond categorizing scene types or localizing object instances, it has been successfully applied to image retrieval~\cite{Johnson-CVPR-Retrieval}, captioning~\cite{li2017scene}, and generation~\cite{johnson2018image}.
Recent advances in scene graph generation leverage the idea of iterative message passing to capture contextual cues and produce a holistic interpretation of the scene~\cite{Qi-ECCV-GraphParsing,Xu-CVPR-MessagePassing,Yang-ECCV-Graph-RCNN,Zellers-CVPR-Neural}. 
%
%
Our work also exploits contextual information but has the following distinctions: 
(1) Unlike existing methods that apply message passing to update \emph{appearance features} (e.g., the appearance feature extracted from the union of human-object pair) at each step, we use an abstract spatial-semantic representation with an \emph{explicit} encoding of relative spatial layout.
(2) In contrast to prior works that use a single densely connected graph structure where edges connecting all possible object pairs,
we operate on human-centric and object-centric subgraphs to focus on relevant contextual information specifically for HOI. 
\figref{motivation} highlights the differences between methods that capture contextual cues. 

The mechanisms for dynamically capturing contextual cues for resolving ambiguity in local predictions have also been successfully applied to sequence prediction~\cite{vaswani2017attention}, object detection~\cite{Hu-CVPR-Relation}, action recognition~\cite{girdhar2018video,Sun-CVPPR-Actor-centric,Wang-CVPR-2018}, and HOI detection~\cite{Gao-BMVC-iCAN}.
Our dual relation graph shares a similar high-level idea with these approaches but with a focus on exploiting the contexts of spatial-semantic representations.

\para{Visual abstraction.} 
The use of visual abstraction helps direct the focus to study the semantics of an image~\cite{zitnick2013bringing}.
Visual abstraction has also been applied to learn common sense~\cite{vedantam2015learning}, forecasting object dynamics~\cite{fouhey2014predicting}, and visual question answering~\cite{yi2018neural}. 
Our work leverages the contexts of an abstract representation between human-object pairs for detecting HOIs.

\para{Spatial-semantic representation.} 
Spatial-semantic representation has also been applied in other problem domains such as image search~\cite{mai2017spatial}, multi-class object detection~\cite{desai2011discriminative}, and image captioning~\cite{yin2017obj2text}.

\section{Method}
\label{sec:overview}
\begin{figure}[t]
\includegraphics[width=\linewidth]{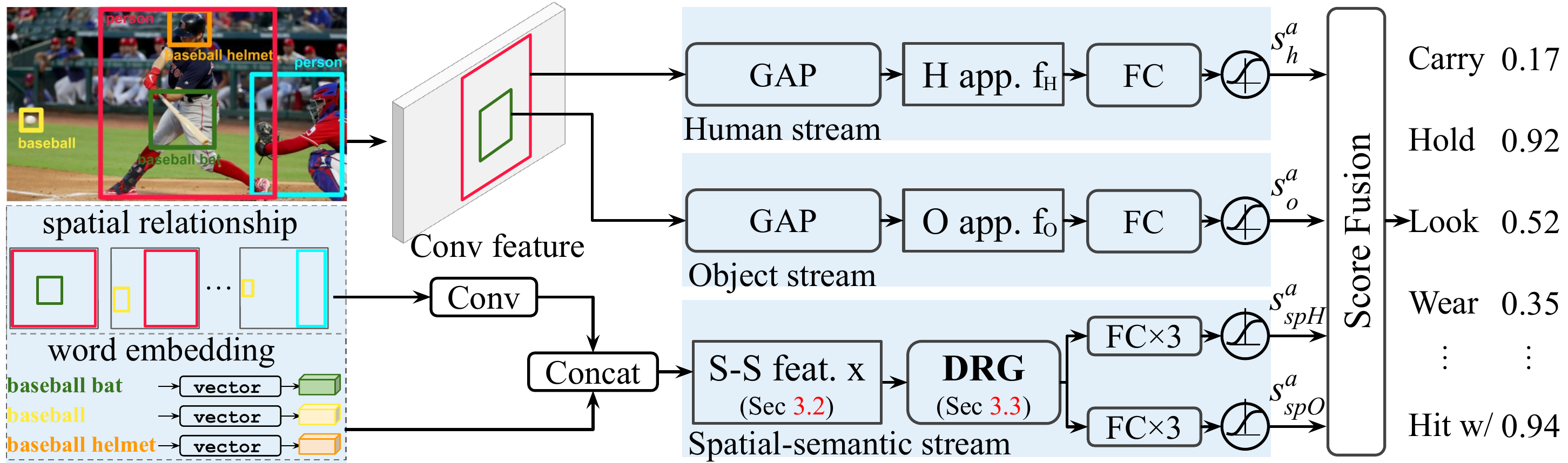}
\caption{\textbf{Overview of the proposed model.} 
Our network consists of three streams (human, object, and spatial-semantic).
The human and object stream leverage the appearance feature $\mathbf{f}_h$ and $\mathbf{f}_o$.
The spatial-semantic stream makes a prediction from the abstract spatial-semantic feature $\bold{x}$. 
We apply our proposed dual relation graph (DRG) to this stream. 
The three streams predict the scores $s_h^a$, $s_o^a$, $s_{spH}^a$ and $s_{spO}^a$, which are fused to form final prediction.
}
\label{fig:overview}
\end{figure}

In this section, we present our network for HOI detection (\figref{overview}).
We start with an overview of our network (\secref{alg_overview}). 
We then introduce the spatial-semantic representation (\secref{s-s_feature}) and describe how we can leverage the proposed Dual Relation Graph (DRG) to propagate contextual information (\secref{RBG}). 
Finally, we outline our inference (\secref{inference}) and the training procedure (\secref{training}).

\subsection{Algorithm overview}
\label{sec:alg_overview}

\figref{overview} provides a high-level overview of our HOI detection network.
We decompose the HOI detection problem into two steps: (1) object detection and (2) HOI prediction.
Following Gao \etal~\cite{Gao-BMVC-iCAN}, we first apply an off-the-shelf object detector Faster R-CNN~\cite{Ren-NIPS-FasterRCNN} to detect all the human/object instances in an image.
We denote $\mathbb{H}$ as the set of human detections, and $\mathbb{O}$ as the set of object detections. 
Note that ``person'' is also an object category.
We denote $b_h$ as the detected bounding box for a person and $b_o$ for an object instance.
We use $s_h$ and $s_o$ to denote the confidence scores produced by the object detector for a detected person $b_h$ and an object $b_o$, respectively. 
Given the detected bounding boxes $b_h$ and $b_o$, we first extract the ROI pooled features and pass them into the human and object stream.
We then pass the detected bounding boxes as well as the object category information to the spatial-semantic stream. 
We apply the proposed \emph{Dual Relation Graph (DRG)} in the spatial-semantic stream.
Lastly, we fuse the action scores from the three streams (human, object, and spatial-semantic) to produce our final predictions.

\para{Human and object stream.} 
%
Our human/object streams follow the standard object detection pipeline for feature extraction and classification.
For each ROI pooled human/object feature, we pass it into a 
one-layer MLP followed by global average pooling and obtain the human appearance feature $\bold{f}_h$ and the object appearance feature $\bold{f}_o$.
We then apply a standard classification layer to obtain the A-dim action scores $s_h^a$ (from human stream) and $s_o^a$ (from object stream).

\para{Spatial-semantic stream.} 
Our inputs to this stream are the spatial-semantic features (described in \secref{s-s_feature}).
In an image, we pair all the detected persons in $\mathbb{H}$ with all the objects in $\mathbb{O}$, and extract spatial-semantic features for each human-object pair.
We then pass all the features into our proposed Dual Relation Graph (\secref{RBG}) to aggregate the contextual information and produce updated features for each human-object pair.
Our dual relation graph consists of a human-centric subgraph and an object-centric subgraph.
These two subgraphs produce the action scores, $s_{spH}^a$ and $s_{spO}^a$.

\begin{figure}[t]
\includegraphics[width=\linewidth]{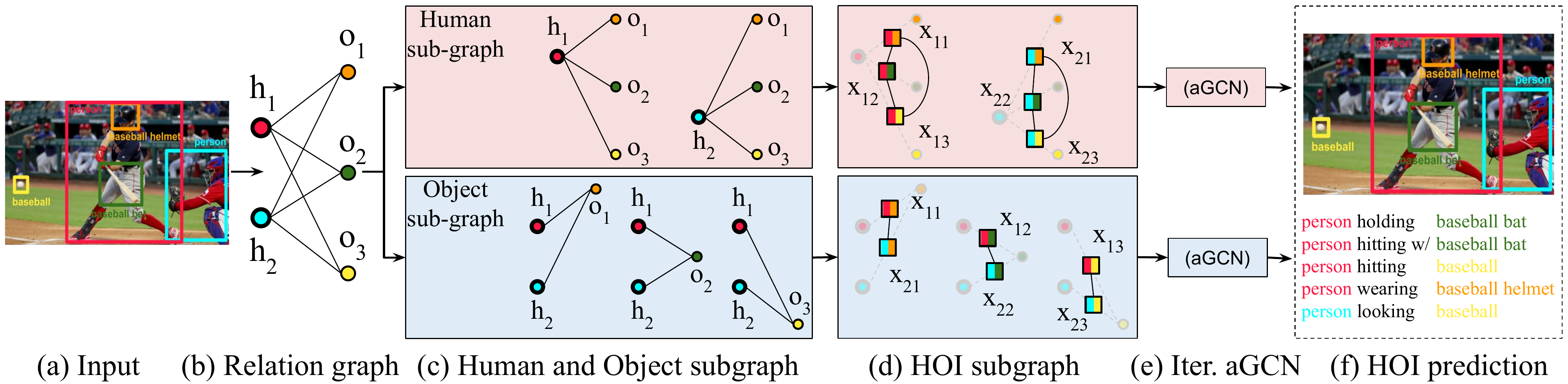}
\caption{\textbf{HOI detection using Dual Relation Graph.} 
(a) The input to our model are the detected objects in the given image. 
We denote $\mathbb{H}$ as the set of human detections, and $\mathbb{O}$ as the set of object detections.
(b) We construct a \emph{relation graph} from the detections where the two sets are  $\mathbb{H}$ and $\mathbb{O}$.
(c) For each human node \textit{h} in $\mathbb{H}$, we obtain a human-centric sub-graph where all nodes in $\mathbb{O}$ are connected to \textit{h}. 
Similarly, we can obtain an object-centric sub-graph for each object node \textit{o} in $\mathbb{O}$. 
Note that ``person'' is also an object category. 
For simplicity, we do not show it in the figure.
(d) In order to predict HOIs, we need to construct the HOI graph explicitly. 
Taking human sub-graph for example, we insert an HOI node \textit{x} between human node \textit{h} and object node \textit{o}. 
We then connect all the HOI nodes and obtain the \textit{human-centric} HOI sub-graph and the \textit{object-centric} HOI sub-graph.
(e) We iteratively update the HOI node feature via a trainable attentional graph convolutional network. 
This helps to aggregate the contextual information. 
(f) We fuse the scores from both sub-graphs and make the final HOI prediction.
}
\label{fig:bipartite_graph}
\end{figure}
\vspace{-4.0mm}
\subsection{Spatial-semantic representation}
\label{sec:s-s_feature}

We leverage the \emph{abstract visual representation} of human-object pair for HOI prediction.
The visual abstraction of a human-object pair allows us to construct representations that are invariant to intra-class appearance variations.
In the context of human-object interaction, we consider the two important visual abstractions: (1) spatial relationship and (2) object category.
\footnote{Other types of abstracted representation such as the pose of the person, the attribute of the person/object can also be incorporated into our formulation. We leave this to future work.}  

\para{Capturing pairwise spatial relationship.} 
Following \cite{Chao-WACV-HOI,Gao-BMVC-iCAN}, we use the two-channel binary image representation to model the spatial relationship between a person and an object.
To do so, we take the union of the two bounding boxes as a reference and rescale it to a fixed size.
A binary image with two channels can then be created by filling value ones within the human bounding box in the first channel and filling value ones within the object bounding box in the second channel.
The remaining locations are then filled with value 0.
We then feed these two-channel binary images into a simple two-layer ConvNet to extract the spatial relation feature.

\para{Capturing object semantics.} 
We find that using spatial features by itself leads to poor results in predicting the interaction. 
To address this issue, we augment the spatial feature with the word embedding of each object's category, $vector$($o$), using fastText~\cite{Mikolov-LREC-fastText}.
Let $\bold{x}_{ij}$ denote the \emph{spatial-semantic feature} between the $i$-th person and the $j$-th object. We construct $\bold{x}_{ij} \in \mathbb{R}^{5708}$ by concatenating (1) the spatial feature and (2) the $ 300$-dimensional word embedding vector.

\subsection{Dual Relation Graph}
\label{sec:RBG}
Here, we introduce the Dual Relation Graph for aggregating the spatial-semantic features.
\figref{bipartite_graph} illustrates the overall process.

\heading{Relation graph.}
Given the object detection results, \ie instances in $\mathbb{H}$ and $\mathbb{O}$, we construct a relation graph (\figref{bipartite_graph}b). 
There are two types of nodes in this graph, $\mathbb{H}$ (human) and $\mathbb{O}$ (object). 
For each node $\textit{h}$ in $\mathbb{H}$, we connect it to all the other nodes in $\mathbb{O}$. 
Similarly, for each node $\textit{o}$ in $\mathbb{O}$, we connect it to all nodes in $\mathbb{H}$.

\heading{Human-centric subgraph and object-centric subgraph.}
Unlike previous methods \cite{Qi-ECCV-GraphParsing, Yang-ECCV-Graph-RCNN}, we do not use the densely connected graphs.
To exploit the relation among different HOIs performed by \emph{the same person}, we construct a \emph{human-centric} subgraph. 
Similarly, we construct an \emph{object-centric} subgraph for the HOIs performed on \emph{the same object} (\figref{bipartite_graph}c).
So far, each node stands for an object instance detection. 
To explicitly represent the HOI, we insert an HOI node $\it{x_{ij}}$ between each paired human node $\it{h_i}$ and object node $\it{o_j}$. 
We then connect all the HOI nodes and obtain \emph{human-centric} HOI subgraph and \emph{object-centric} HOI subgraph (\figref{bipartite_graph}d). 
We use the before mentioned \emph{spatial-semantic feature} $\bold{x}_{ij}$ to encode each HOI node between the $i$-th person and the $j$-th object.


\heading{Contextual feature aggregation.}
With these two HOI subgraphs, we follow a similar procedure for propagating and aggregating features as in relation network~\cite{Hu-CVPR-Relation}, non-local neural network~\cite{Wang-CVPR-2018}, and attentional graph convolutional network~\cite{Yang-ECCV-Graph-RCNN}.

\emph{Human-centric HOI subgraph}. To update node $\it{x_{ij}}$, we aggregate all the spatial-semantic feature of the nodes involving the same $i$-th person $\{x_{ij^\prime}| j^\prime \in \mathcal{N}(j)\}$. 
%
%
The feature aggregation can be written as:
\begin{equation} \label{eq:feature_updating}
\bold{x}_{ij}^{(l+1)} = \sigma \left( \bold{x}_{ij}^{(l)} + \sum_{j^\prime \in \mathcal{N}(j)} \alpha_{j j^\prime} W \bold{x}_{ij^\prime}^{(l)}\right), 
\end{equation}
where $W \in \mathbb{R}^{5708 \times 5708}$ is a learned linear transformation that projects features into the embedding space, $\alpha_{i j^\prime}$ is a learned attention weight. 

We can rewrite this equation compactly with matrix operation: 
%
\begin{align} \label{eq:feature_updating_matrix}
\bold{x}_{ij}^{(l+1)} & = \sigma \left( W X^{(l)} \bm{\alpha}_j\right) \\
u_{jj^\prime} & = \left(W_q \bold{x}_{ij^\prime}^{(l)}\right)^\top \left(W_k \bold{x}_{ij}^{(l)} \right) / \sqrt{d_k} \\
\bm{\alpha}_j & = \text{softmax} (\bm{u}_{j}),
\end{align}
where $W_q, W_k \in \mathbb{R}^{1024 \times 5708}$ are linear projections that project feature into a query and a key embedding space. 
Following~\cite{vaswani2017attention}, we calculate the attention weights using scaled dot-product, normalized by $\sqrt{d_k}$ where $d_k = 1024$ is the dimension of the key embedding space.
We do not directly use the aggregated feature $\sigma \left( W X^{(l)} \bm{\alpha}_j\right)$ as our output updated feature. Instead, we add it back to the original spatial-semantic feature $\bold{x}_{ij}^{(l)}$.
We then pass the addition through a LayerNorm to get the final aggregated feature on the human-centric subgraph.
\begin{equation} \label{eq:updating}
\bold{x}_{ij}^{(l+1)} = \text{LayerNorm}\left(\bold{x}_{ij}^{(l)} + \sigma \left( W X^{(l)} \bm{\alpha}_j\right) \right).
\end{equation}

The linear transformation $W$ does not change the dimension of the input feature, thus the output $\bold{x}_{ij}^{(l+1)}$ has the same size as input $\bold{x}_{ij}^{(l)}$.
As a result, we can perform several iterations of feature aggregation (\figref{bipartite_graph}e). 
We explore the effectiveness of more iteration in \tabref{ablation}(a).\\

\emph{Object-centric HOI subgraph}. Similarly, to update node $\it{x_{ij}}$, we aggregate all the spatial-semantic feature of the nodes which involved the same object $\{x_{i^\prime j}| i^\prime \in \mathcal{N}(i)\}$. 
The two subgraphs have independent weights and aggregate contextual information independently.

\subsection{Inference}
\label{sec:inference}
%
%
For each human-object bounding box pair ($b_h$, $b_o$) in image $I$, we predict the score $S^a_{h,o}$ for each action $a \in \left\{ 1, \cdots, A \right\}$, where $A$ denotes the total number of possible actions.
The final score $S^a_{h,o}$ depends on  
(1) the confidence for the individual object detections ($s_h$ and $s_o$), 
(2) the prediction score from the appearance of the person $s^a_h$ and the object $s^a_o$, and 
(3) the prediction score based on the aggregated spatial-semantic feature, using human-centric and object-centric subgraph, $s_{spH}^a$ and $s_{spO}^a$.
We compute the HOI score $S^a_{h,o}$ for the human-object pair ($b_h$, $b_o$) as
\begin{align}    
S^a_{h,o} & = s_h \cdot s_o \cdot s^a_{h} \cdot s^a_{o} \cdot s^a_{spH} \cdot s^a_{spO}
\end{align}
Note that we are not able to obtain the action scores using object $s^a_{o}$ or the spatial-semantic stream for some classes of actions as they do not involve any objects (\eg walk, smile).
For those cases, we use only the score $s^a_h$ from the human stream.
For those actions, our final scores are $s_h \cdot s^a_h$.

\subsection{Training}
\label{sec:training}
HOI detection is a multi-label classification problem because a person can simultaneously perform different actions on different objects, e.g., sitting on a chair and reading a book.
Thus, we minimize the cross-entropy loss \emph{for each individual action class} between the ground-truth action label and the score produced from each stream. 
The total loss is the summation of the loss at each stream.
\section{Experimental Results}
\label{sec:results}
\newlength\ftqa
\setlength\ftqa{2.0cm}
\newlength\ftqb
\setlength\ftqb{0.2mm}
\newlength\ftqc
\setlength\ftqc{0.8mm}

\begin{figure*}[t]

\includegraphics[width=\ftqa]{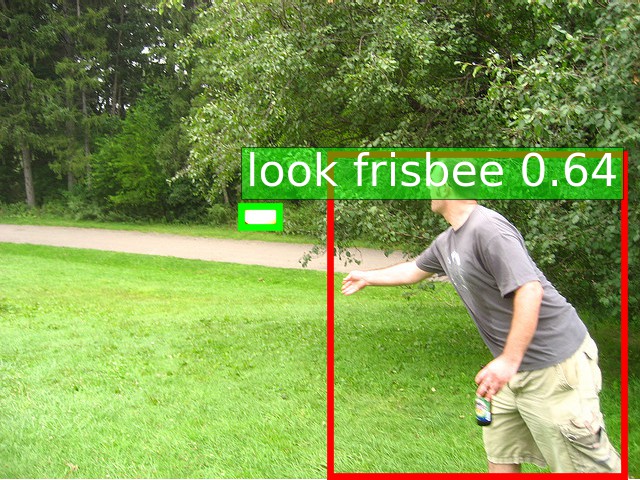}\hfill%
\includegraphics[width=\ftqa]{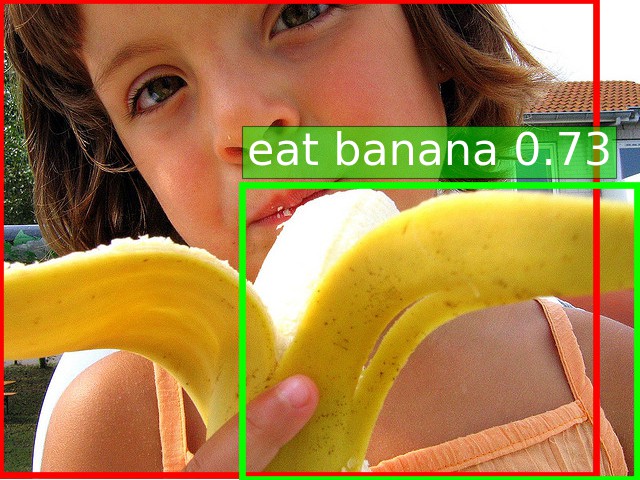}\hfill%
\includegraphics[width=\ftqa]{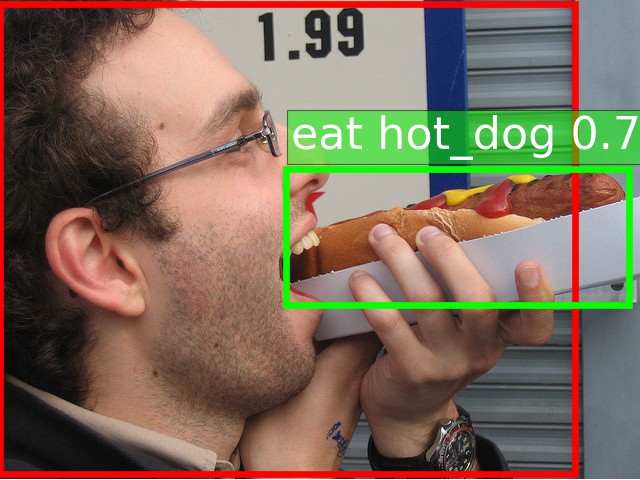}\hfill%
\includegraphics[width=\ftqa]{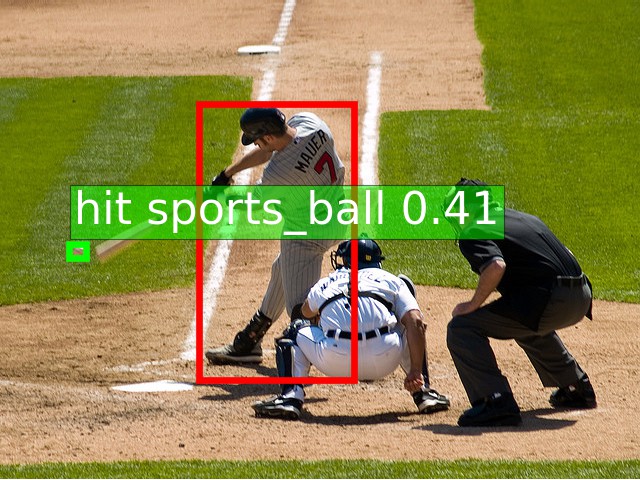}\hfill%
\includegraphics[width=\ftqa]{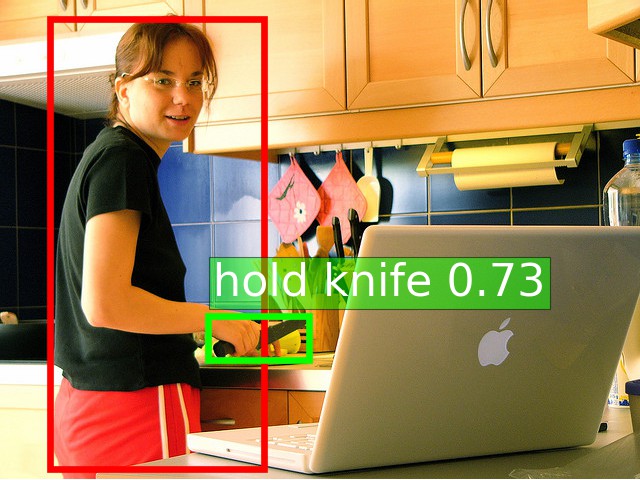}\hfill%
\includegraphics[width=\ftqa]{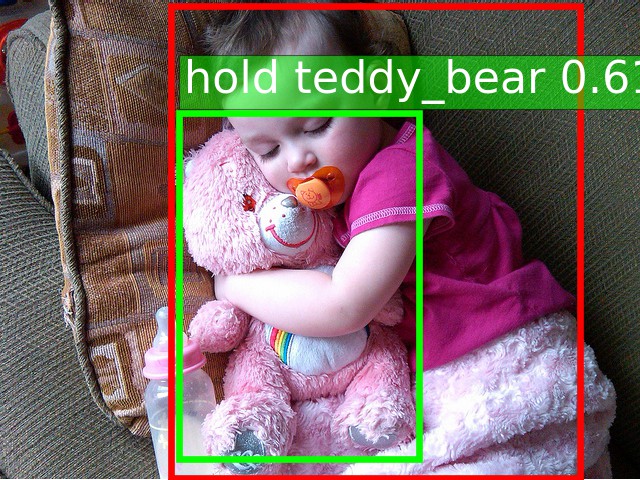}\vspace{\ftqb}\\%
\includegraphics[width=\ftqa]{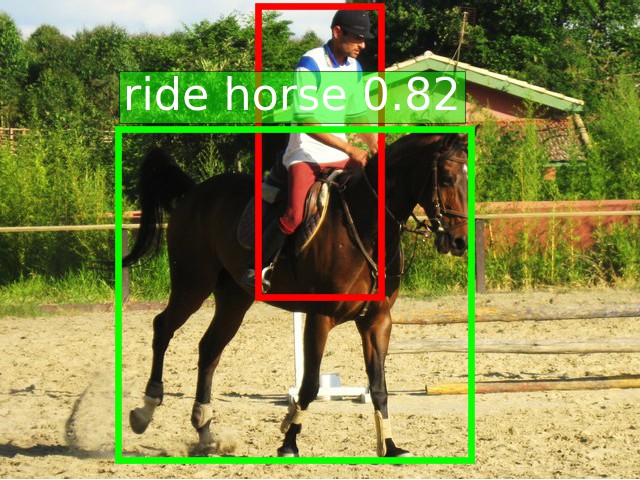}\hfill%
\includegraphics[width=\ftqa]{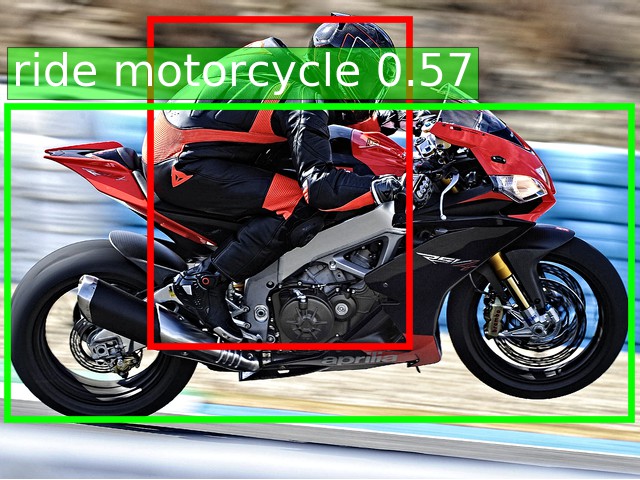}\hfill%
\includegraphics[width=\ftqa]{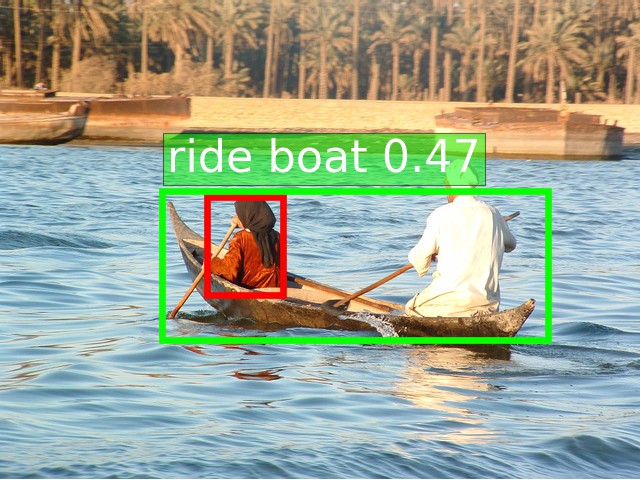}\hfill%
\includegraphics[width=\ftqa]{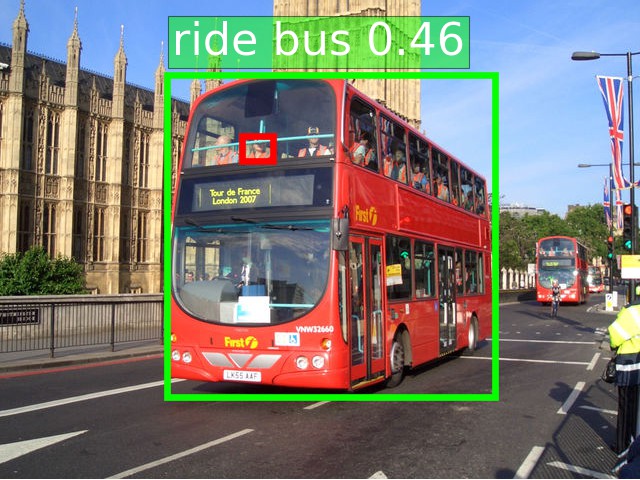}\hfill%
\includegraphics[width=\ftqa]{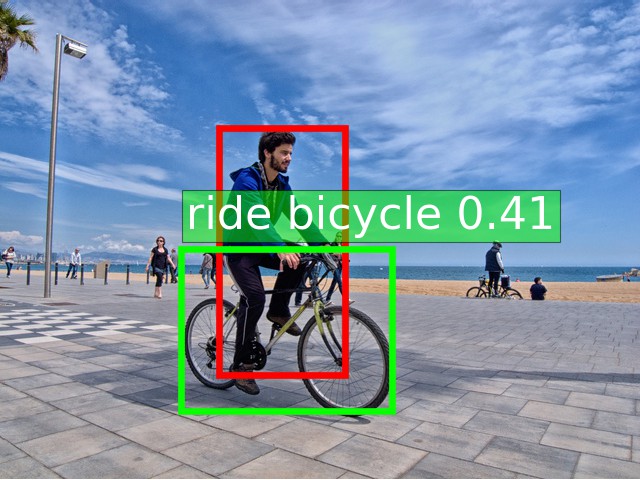}\hfill%
\includegraphics[width=\ftqa]{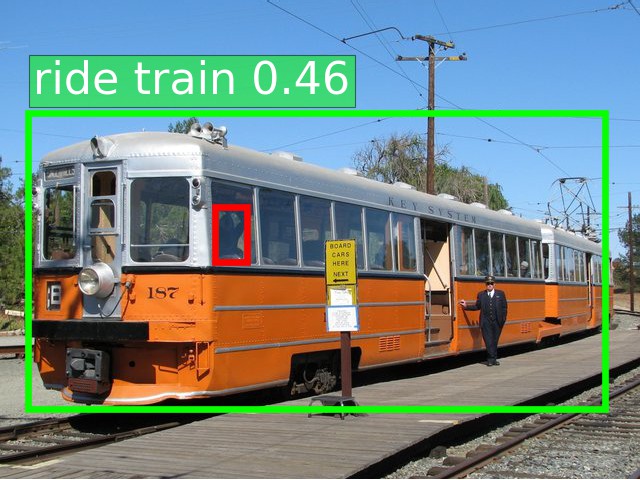}
\caption{\textbf{Sample HOI detections on V-COCO (first row) and HICO-DET (second row) \emph{test} set.}
\label{fig:visual}
}
\end{figure*}
\vspace{-3.0mm}

In this section, we first outline our experimental setup, including datasets, metrics, and implementation details. 
We then report the quantitative results on two large-scale HOI benchmark datasets and compare the performance with the state-of-the-art HOI detection algorithms.
Next, we show sample visual results on HOI detection. 
We conduct a detailed ablation study to quantify the contributions from individual components and validate our design choices.
More results can be found in the supplementary material.
We will make the source code and pre-trained models publicly available to foster future research.

\begin{table}[t]
\centering
\caption{\tb{Comparison with the state-of-the-art on the V-COCO $\emph{test}$ set.} The best performance is in \first{bold} and the second best is \second{underscored}. Character $^\star$ indicates that the method uses both VCOCO and HICO-DET training data.
``S-S only'' shows the performance of our spatial-semantic stream.
}
\label{tab:vcoco_comparison}
{
\begin{tabular}{lccl|c}
\toprule
Method & Use human pose &&  Feature backbone & \textbf{$AP_{role}$} \\
\midrule
VSRL~\cite{Gupta-SemanticRoleLabeling}              & - && ResNet-50-FPN    & 31.8 \\
InteractNet~\cite{Gkioxari-CVPR-InteractNet}        & - && ResNet-50-FPN    & 40.0 \\
BAR-CNN~\cite{Kolesnikov-BAR}                       & - && Inception-ResNet & 41.1 \\
GPNN~\cite{Qi-ECCV-GraphParsing}                    & - && ResNet-101       & 44.0 \\
iCAN~\cite{Gao-BMVC-iCAN}                           & - && ResNet-50        & 45.3 \\
Wang et al. ~\cite{wang2019deep} & - && ResNet-50 & 47.3 \\
RPNN~\cite{zhou2019relation} & \checkmark && ResNet-50 & 47.5 \\
${RP_{T2}C_D}^\star$~\cite{Li-CVPR-Interactiveness} & \checkmark && ResNet-50        & 48.7 \\
PMFNet \cite{Bo-PMFNet}                             & - && ResNet-50-FPN    & 48.6 \\
PMFNet \cite{Bo-PMFNet}                             & \checkmark && ResNet-50-FPN    & \first{52.0} \\
Ours (S-S only)                               & - & & -           & 47.1 \\
Ours                                                & - && ResNet-50-FPN    & \second{51.0} \\
\bottomrule
\end{tabular}
}
\end{table}

\subsection{Experimental setup}
\para{Datasets.}
%
\tb{V-COCO dataset}~\cite{Gupta-SemanticRoleLabeling} is constructed by augmenting the COCO dataset~\cite{Lin-ECCV-MSCOCO} with additional human-object interaction annotations.
%
%
Each person is annotated with a binary label vector for 29 different action categories (five of them do not involve associated objects).
\tb{HICO-DET}~\cite{Chao-CVPR-HICO} is a larger dataset containing 600 HOI categories over 80 object categories (same as \cite{Lin-ECCV-MSCOCO}) with more than 150K annotated instances of human-object pairs.
%
For applying our method on the HICO-DET dataset, we disentangle the 600 HOI categories into 117 object-agnostic action categories and train our network over these 117 action categories.
At test time, we then combine the predicted action and the detected object and convert them back to the original 600 HOI classes.
Note that the evaluation for the HICO-DET dataset remains the same.
%

\para{Evaluation metrics.}
To evaluate the performance of our model, we adopt the commonly used role mean average precision (role mAP)~\cite{Gupta-SemanticRoleLabeling} for both V-COCO and HICO datasets.
%
The goal is to detect and correctly predict the $\langle$ $\texttt{human, verb, object}$ $\rangle$ triplet.
%
We consider a triplet as true positive if and only if it localizes the human and object accurately (\ie with IoUs $\geq 0.5$ w.r.t the ground truth annotations) and predicts the action correctly.

\begin{table*}[t]
\centering
\caption{\tb{Comparison with the state-of-the-art on HICO-DET \emph{test} set.} The best performance is in \first{bold} and the second best is \second{underscored}. Character $^\star$ indicates that the method uses both VCOCO and HICO-DET training data.
For the object detector, ``COCO'' means that the detector is trained on COCO, while ``HICO-DET'' means that the detector is first pre-trained on COCO and then further fine-tuned on HICO-DET.
}
\label{tab:HICO}
\resizebox{\columnwidth}{!}{

\begin{tabular}{l cc l ccc c  ccc}
\toprule
\multirow{4}{*}{} & &&&
\multicolumn{3}{c}{Default} & &
\multicolumn{3}{c}{Known Object}\\
\cline{5-7} \cline{9-11}
Method & Detector & Use human pose & Feature backbone & Full & Rare & Non Rare &  & Full & Rare & Non Rare  \\
\midrule
Shen~\etal~\cite{Shen-WACV-Zeroshot} & COCO & - & VGG-19 & 6.46 & 4.24 & 7.12 & & - & - & -\\
HO-RCNN~\cite{Chao-WACV-HOI} & COCO & - & CaffeNet & 7.81 & 5.37 & 8.54 & & 10.41 & 8.94 & 10.85\\
InteractNet~\cite{Gkioxari-CVPR-InteractNet} & COCO & - & ResNet-50-FPN & 9.94 & 7.16 & 10.77 & & - & - & -\\
GPNN~\cite{Qi-ECCV-GraphParsing} & COCO & - & ResNet-101 & 13.11 & 9.34 & 14.23 &  & - & - & - \\
iCAN~\cite{Gao-BMVC-iCAN} & COCO & - & ResNet-50 & 14.84 & 10.45 & 16.15 & & 16.26 & 11.33 & 17.73 \\
Wang et al.~\cite{wang2019deep} & COCO & - & ResNet-50 & 16.24 & 11.16 & 17.75 && 17.73 & 12.78 & 19.21 \\
Bansal~\etal~\cite{bansal2019detecting} & COCO & - &  ResNet-101 & 16.96 & 11.73 & 18.52 &  & - & - & - \\
$RP_DC_D$~\cite{Li-CVPR-Interactiveness} & COCO & \checkmark & ResNet-50 & 17.03 & 13.42 & 18.11 &  & 19.17 & 15.51 & 20.26 \\
${RP_{T2}C_D}^\star$~\cite{Li-CVPR-Interactiveness} & COCO & \checkmark & ResNet-50 & 17.22 & 13.51 & 18.32 &  & 19.38 & 15.38 & 20.57 \\
no-frills~\cite{Alex-No-Frills} & COCO & \checkmark & ResNet-152 & 17.18 & 12.17 & 18.68 && - & - & -\\
RPNN~\cite{zhou2019relation} & COCO & \checkmark & ResNet-50 & 17.35 & 12.78 & 18.71 & &-&- & - \\
PMFNet~\cite{Bo-PMFNet} & COCO & - & ResNet-50-FPN & 14.92 & 11.42 & 15.96 & & 18.83 & 15.30 & 19.89 \\
PMFNet~\cite{Bo-PMFNet} & COCO & \checkmark & ResNet-50-FPN & 17.46 & \second{15.65} & 18.00 & & \second{20.34} & \second{17.47} & \second{21.20} \\
Peyre~\etal~\cite{peyre2018detecting} & COCO & - & ResNet-50-FPN & \first{19.40} & 14.63 & \first{20.87} & & - & - & - \\
Ours (S-S only)  & COCO & -   & -      & 12.45 & 9.84  & 13.23 & & 15.77 & 12.76 & 16.66 \\
Ours & COCO & - &  ResNet-50-FPN & \second{19.26} & \first{17.74} & \second{19.71} & & \first{23.40} & \first{21.75} & \first{23.89} \\ 
\midrule
Bansal~\etal~\cite{bansal2019detecting} & HICO-DET & - &  ResNet-101 & \second{21.96} & \second{16.43} & \second{23.62} &  & - & - & - \\
Ours & HICO-DET & - &  ResNet-50-FPN & \first{24.53} & \first{19.47} & \first{26.04} &  & 27.98 & 23.11 & 29.43 \\
\bottomrule
\end{tabular}

}
\end{table*}

\para{Implementation details.} 
%
We build our network with the publicly available PyTorch framework.
Following Gao~\etal~\cite{Gao-BMVC-iCAN}, we use the Detectron \cite{Detectron} with a feature backbone of ResNet-50 to generate human and object bounding boxes.
For VCOCO, we conduct an ablation study on the validation split to determine the best threshold. We keep the detected human boxes with scores $s_h$ higher than 0.8 and object boxes with scores $s_o$ higher than 0.1. 
For HICO-DET, since there is no validation split available, we follow the setting in \cite{peyre2018detecting}. We use the score threshold 0.6 to filter out unreliable human boxes and threshold 0.4 to filter out unconfident object boxes.
To augment the training data, we apply random spatial jitterring to the human and object bounding boxes and ensure that the IOU with the ground truth bounding box is greater than 0.7.
We pair all the detected human and objects, and regard those who are not ground truth as negative training examples. 
We keep the negative to positive ratio to three.

We initialize our appearance feature backbone with the COCO pre-trained weight from Mask R-CNN \cite{He-ICCV-MaskRCNN}.
We perform two iterations of feature aggregation on both \emph{human-centric} and \emph{object-centric} subgraphs. 
We train the three streams (human appearance, object appearance, and spatial-semantic) using the V-COCO \emph{train} set.
We use early stopping criteria by monitoring the validation loss.
We train our network with a learning rate of $0.0025$, a weight decay of $0.0001$, and a momentum of $0.9$ on both the V-COCO \emph{train} set and HICO-DET \emph{train} set. 
Training our network takes $14$ hours on a single NVIDIA P100 GPU on V-COCO and $24$ hours on HICO-DET.
%
%
At test time, our model runs at $3.3$ fps for VCOCO and $5$ fps for HICO-DET.


\subsection{Quantitative evaluation}

We report the main quantitative results in terms of $AP_{role}$ on V-COCO in \tabref{vcoco_comparison} and HICO-DET in \tabref{HICO}.
For the V-COCO dataset, our method compares favorably against state-of-the-art algorithms~\cite{Li-CVPR-Interactiveness,wang2019deep,zhou2019relation} except PMFNet \cite{Bo-PMFNet}, which uses human pose as an additional feature.
Since pose estimation required additional training data (with pose annotations), we expect to see performance gain using human pose.
PMFNet \cite{Bo-PMFNet} also reports the $AP_{role}$ \emph{without} human pose, which is $2.4$ mAP lower to our method. 
We also note that the spatial-semantic stream alone \emph{without} using any visual features achieves a competitive performance ($47.1$ mAP) when compared with the state-of-the-art. 
This highlights the effectiveness of the abstract spatial-semantic representation and contextual information.
Compared with methods that perform joint inference on densely connected graph~\cite{Qi-ECCV-GraphParsing},
our approach produces significant performance gains.

For the HICO-DET dataset, our method also achieves competitive performance with state-of-the-art methods~\cite{Alex-No-Frills,Li-CVPR-Interactiveness,Bo-PMFNet,Peyre-ICCV-Weakly}.
Our method achieves the best performance for the \emph{rare categories}, showing that our method handles the long-tailed distributions of HOI classes well.

We note that the current best performing model~\cite{bansal2019detecting} uses an object detector which is fine-tuned on HICO-DET \emph{train} set using the annotated object bounding boxes.
For a fair comparison, we also fine-tune our object detector on HICO-DET and report our result.
Note that we do \emph{not} re-train our model, but only replace the object detector at the test time.

Here, the large performance gain from fine-tuning the object detector may not reflect the progress on the HOI detection task.
This is because the objects (and the associated HOIs) in the HICO-DET dataset are \emph{not} fully annotated. 
Using a fine-tuned object detector can thus improve the HOI detection performance by exploiting such annotation biases. 
\vspace{-2.0mm}
\subsection{Qualitative evaluation}


\para{HOI detection results.}
Here we show sample results on the V-COCO dataset and the HICO-DET dataset in \figref{visual}. 
We highlight the detected person and the associated object with red and green bounding boxes, respectively.
%

\newlength\ftqd
\setlength\ftqd{1.9cm}
\begin{figure*}[t]
\centering

\mpage{0.01}{\rotatebox[origin=c]{90}{\small Human-centric}}
\mpage{0.95}{
\includegraphics[width=\ftqd]{figure_SSAR/human_000000235316.png}\hfill%
\includegraphics[width=\ftqd]{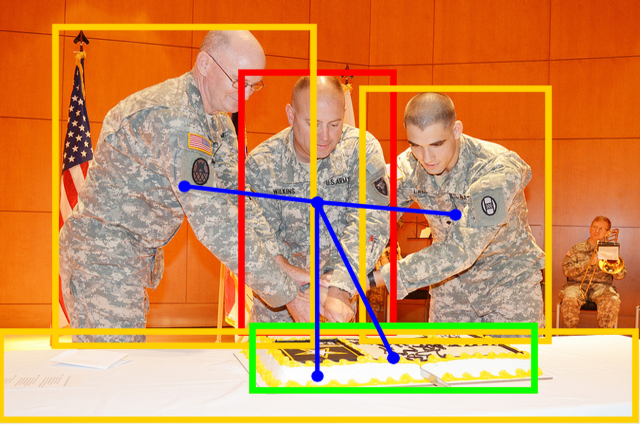}\hfill%
\includegraphics[width=\ftqd]{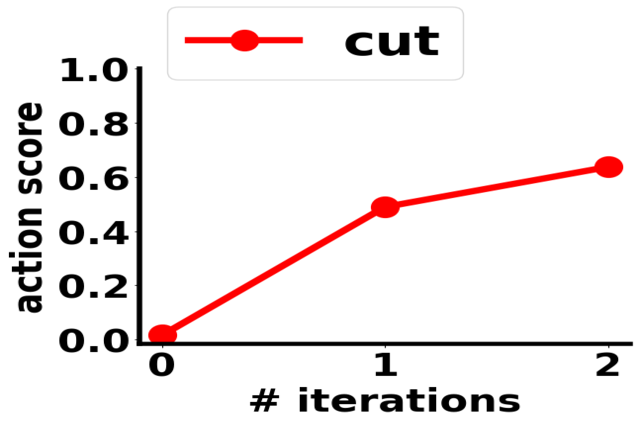}\hfill%
\includegraphics[width=\ftqd]{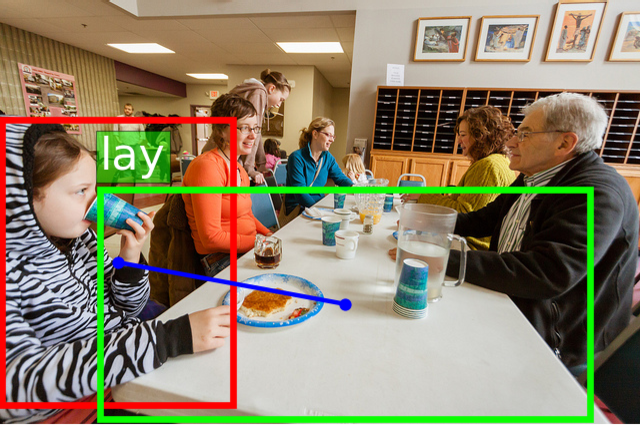}\hfill%
\includegraphics[width=\ftqd]{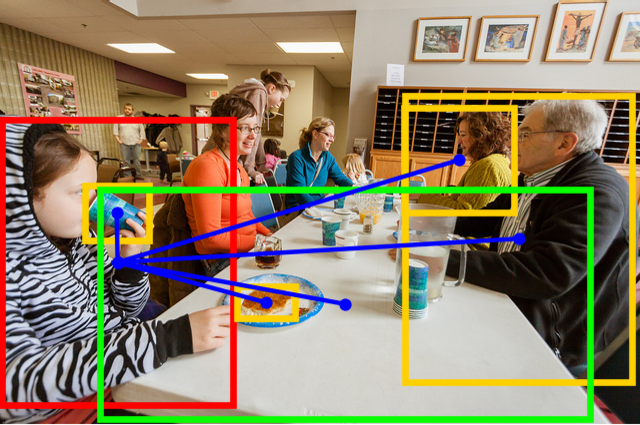}\hfill%
\includegraphics[width=\ftqd]{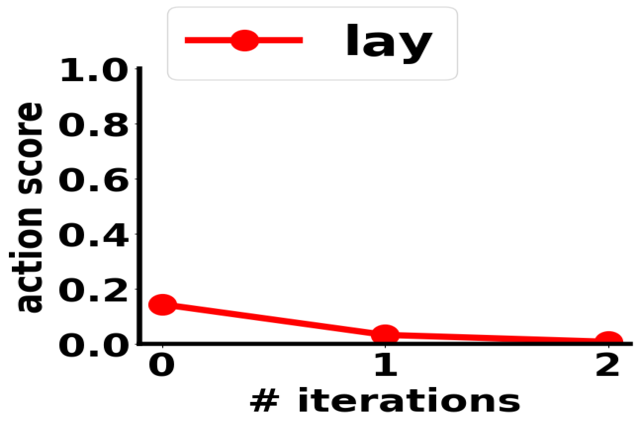}\hfill
\\
\includegraphics[width=\ftqd]{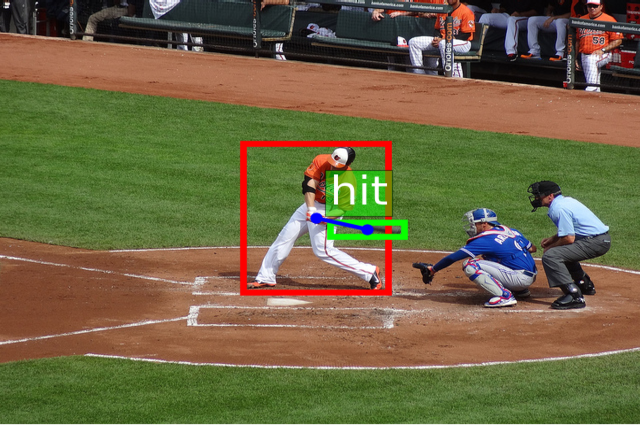}\hfill%
\includegraphics[width=\ftqd]{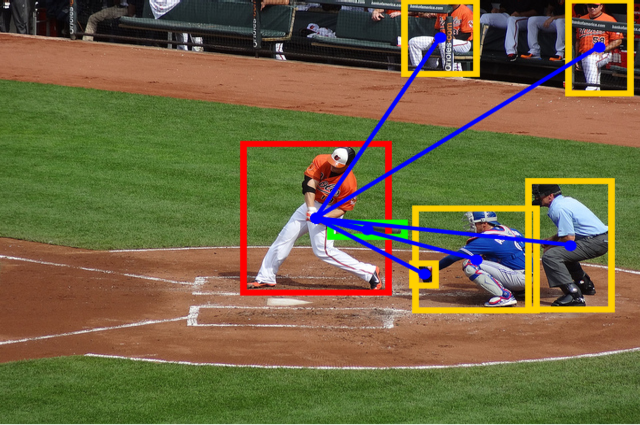}\hfill%
\includegraphics[width=\ftqd]{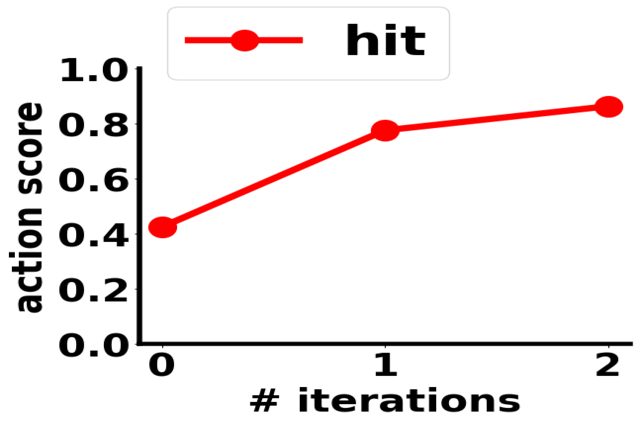}\hfill%
\includegraphics[width=\ftqd]{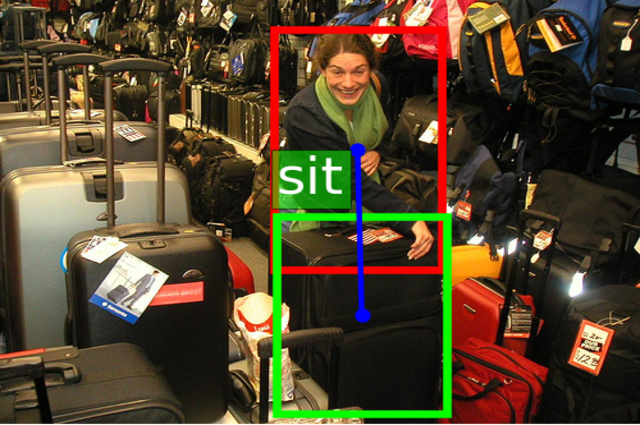}\hfill%
\includegraphics[width=\ftqd]{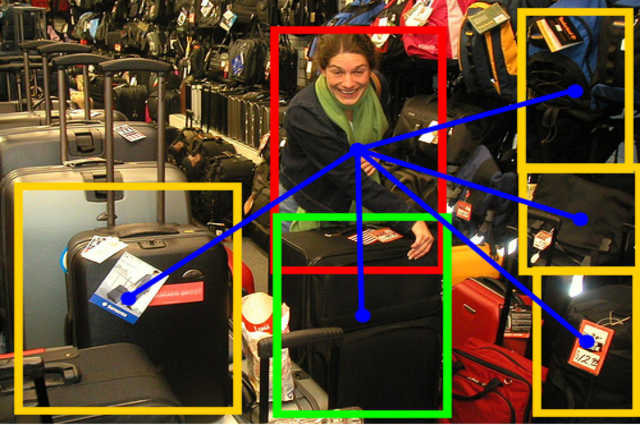}\hfill%
\includegraphics[width=\ftqd]{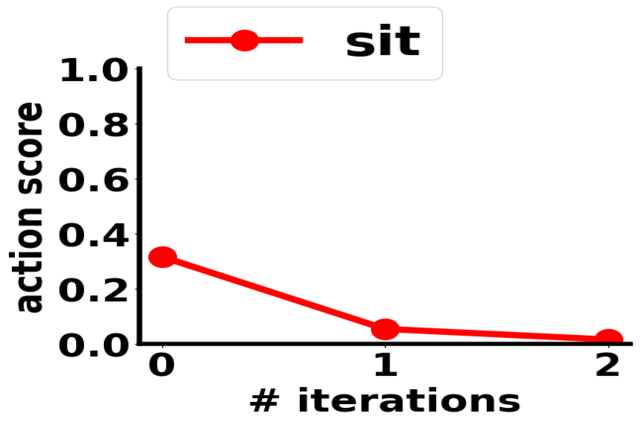}\hfill
\\
}

\mpage{0.01}{\rotatebox[origin=c]{90}{\small Object-centric}}
\mpage{0.95}{
\includegraphics[width=\ftqd]{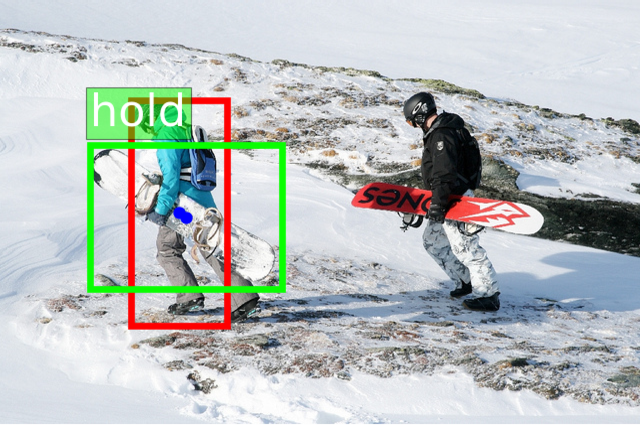}\hfill%
\includegraphics[width=\ftqd]{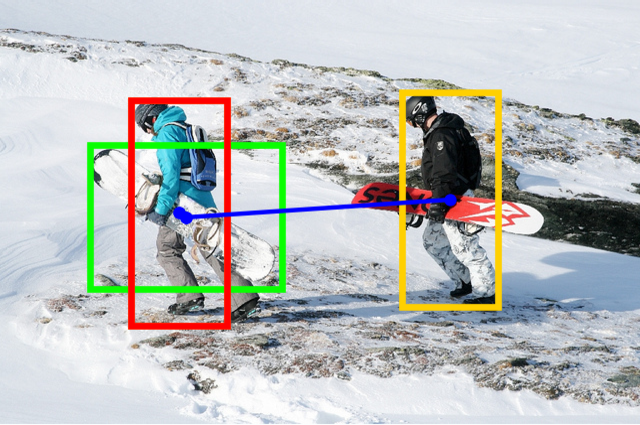}\hfill%
\includegraphics[width=\ftqd]{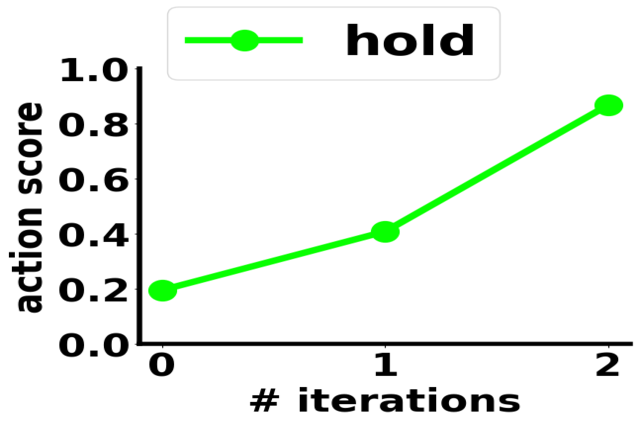}\hfill%
\includegraphics[width=\ftqd]{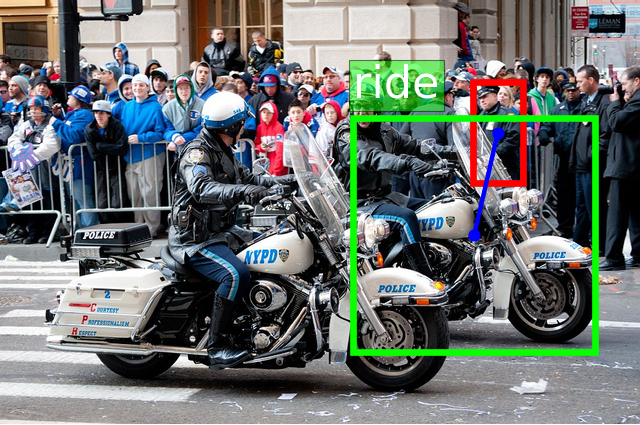}\hfill%
\includegraphics[width=\ftqd]{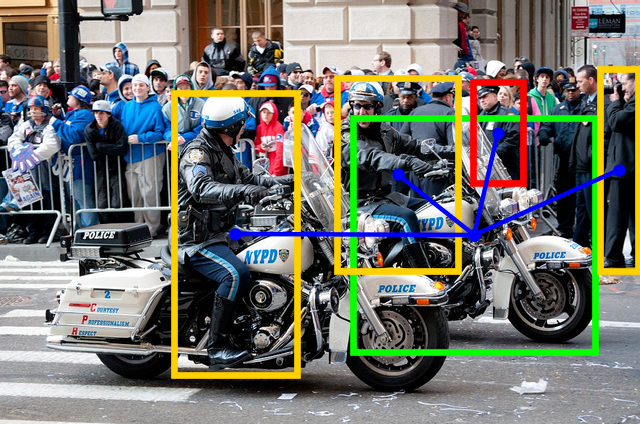}\hfill%
\includegraphics[width=\ftqd]{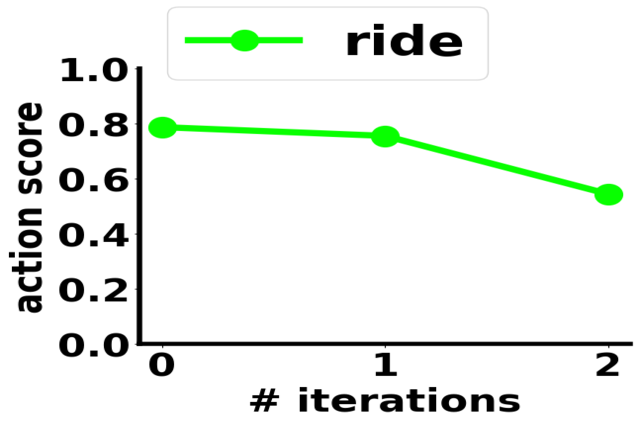}\hfill
\\
\includegraphics[width=\ftqd]{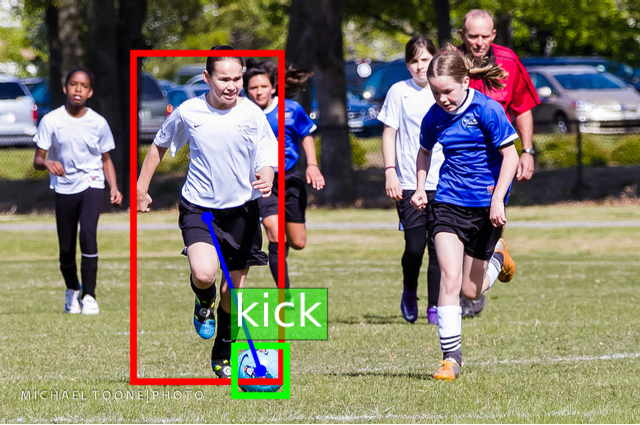}\hfill%
\includegraphics[width=\ftqd]{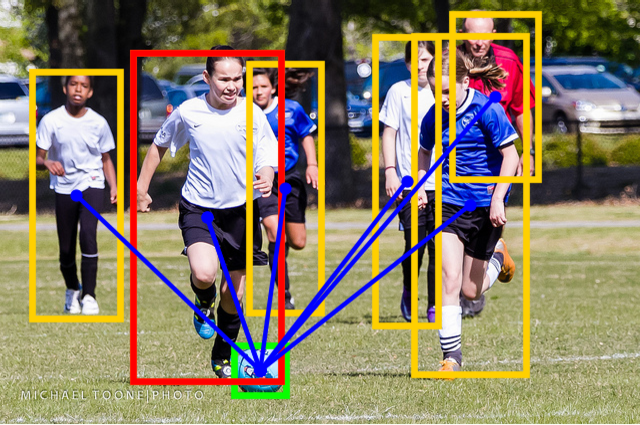}\hfill%
\includegraphics[width=\ftqd]{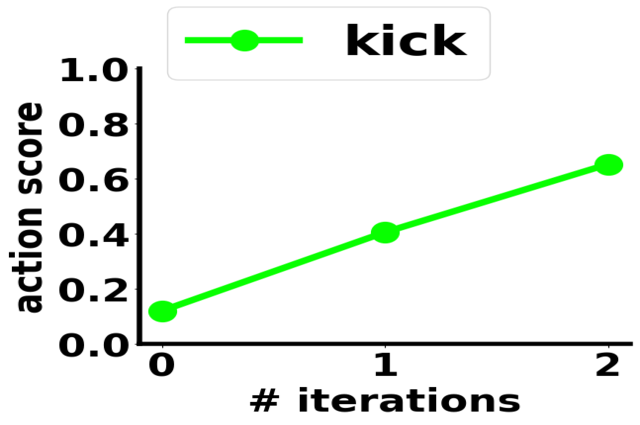}\hfill%
\includegraphics[width=\ftqd]{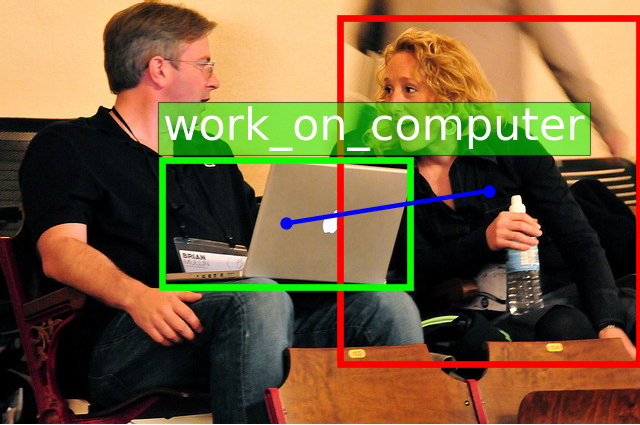}\hfill%
\includegraphics[width=\ftqd]{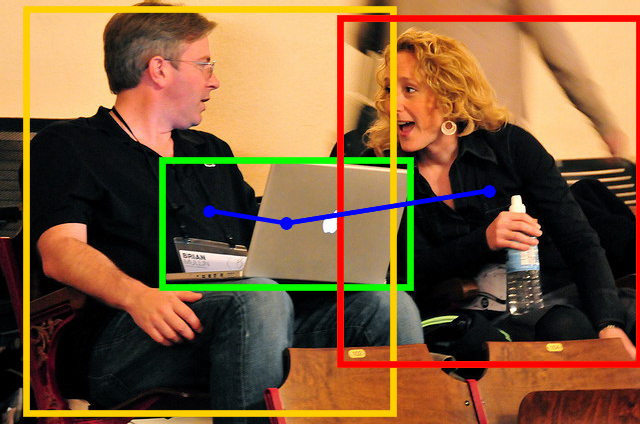}\hfill%
\includegraphics[width=\ftqd]{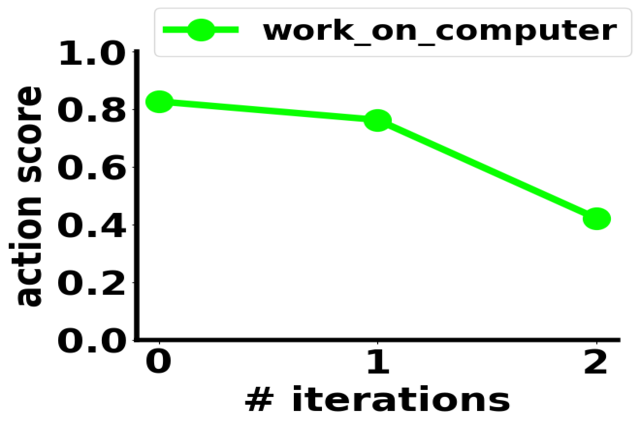}\hfill
\\
}
%

\mpage{0.04}\hfill
\mpage{0.13}{HOI det.} \hfill
\mpage{0.13}{H-O pair} \hfill
\mpage{0.13}{Score} \hfill
\mpage{0.13}{HOI det.} \hfill
\mpage{0.14}{H-O pair} \hfill
\mpage{0.13}{Score}\hfill
\\
\mpage{0.015}\hfill
\mpage{0.46}{
$\underbracket[1pt][2.0mm]{\hspace{\textwidth}}_
    {\substack{\vspace{-1.0mm}\\\colorbox{white}{~~Increasing scores for correct HOIs~~}}}$
}\hfill
\mpage{0.485}{
$\underbracket[1pt][2.0mm]{\hspace{\textwidth}}_
    {\substack{\vspace{-1.0mm}\\\colorbox{white}{Suppressing scores for unrelated pairs}}}$
}\hfill
\caption{\textbf{More iteration of feature aggregation leads to a more accurate prediction.} 
The human-centric and object-centric subgraph in the spatial-semantic stream propagates contextual information to produce increasingly accurate HOI predictions.
}
\label{fig:RBG}
\end{figure*}

\vspace{1.0mm}
\begin{table*}[htbp]
\caption{\textbf{Ablation study on the V-COCO \emph{val} set.} We show the role mAP $AP_{role}$.
}
\label{tab:ablation}

\centering
\mpage{0.5}{
(a) \textbf{More message passing iters.}
} 
\hfill
\mpage{0.45}{
(b) \textbf{Feature used in DRG} 
}
\hfill
\mpage{0.45}{
\begin{tabular}{c|c|c|c}
 iter.    & H graph & O graph & H  + O \\
\hline
 0-iter.  &  48.78 & 47.47 & 50.14 \\
 1-iter.  &  48.83 & 47.35 & 50.74 \\
 2-iter.  &  50.20 & 47.87 & 51.37 \\
\end{tabular}
}
\hfill 
\mpage{0.5}{
\begin{tabular}{l|c}
 & mAP \\
\hline
App. feature (entire image)            & 35.69 \\
App. feature (H-O union box)           & 46.93 \\
Word2vec embedding                           & 37.36 \\
Spatial-semantic feature (ours)              & 51.37 \\
\end{tabular}
}
\hfill 
\\
\mpage{0.35}{
\centering
(c) \textbf{Different subgraph}
} 
\hfill
\mpage{0.6}{
\centering
(d) \textbf{Effectiveness of O subgraph} 
}
\hfill
\mpage{0.35}{
\centering
\begin{tabular}[t]{cc|c}
H graph     & O graph    & mAP   \\
\hline
      -     &      -     & 50.14 \\
 \checkmark &      -     & 51.10 \\
      -     & \checkmark & 50.78 \\
\checkmark & \checkmark  & 51.37 \\
\end{tabular}
}
\hfill
\mpage{0.6}{
\centering
\begin{tabular}[t]{l|cccc}
                             & 1-3 & 4-6 & 7+ & all  \\
\hline
H graph                      & 57.89 & 52.77 & 50.96 & 51.10 \\
H graph + O graph            & 58.28 & 53.75 & 51.06 & 51.37 \\
Margin                       & +0.39 & +0.98 & +0.10 & +0.27 \\
\% of testing images         & 68\%  & 12\%  & 20\%  & 100\% \\
\end{tabular}
}
\end{table*}

\vspace{-2.0mm}

\para{Visualizing the effect of the Dual Relation Graph.} 
In \figref{RBG}, we show the effectiveness of the proposed DRG. 
In the first two rows, we show that by aggregating contextual information, using the  \emph{human-centric} subgraph produces more accurate HOI predictions.
%
Another example in the top right image indicates that the \emph{human-centric} subgraph can also suppress the scores for unrelated human-object pairs. 
%
In the bottom two rows, we show four examples of how the \emph{object-centric} subgraph propagates contextual information in each step to produce increasingly more accurate HOI predictions.
For instance, the bottom right images show that for a person and an object without interaction, our model learns to suppress the predicted score by learning from the relationship of other HOI pairs associated with this particular object (laptop). 
In this example, the model starts with predicting a high score for the woman working on a computer. 
By learning from the relationship between the man and the computer, our model suppresses the score in each iteration.


%

\subsection{Ablation study}
\label{sec:ablation}
We examine several design choices using the V-COCO \emph{val} set.

\para{More iteration of feature aggregation.} 
\tabref{ablation}(a) shows the performance using different iterations of feature aggregation.
For either \emph{human-centric} or \emph{object-centric} subgraph, using more iterations of feature aggregation improves the overall performance.
This highlights the advantages of exploiting contextual information among different HOIs.
Performing feature aggregation on \emph{both} subgraphs further improves the final performance.

\para{Effectiveness of each subgraph.}
To validate the effectiveness of the proposed subgraph, we show different variants of our model in~\tabref{ablation}(c).
Adding only \emph{human-centric} subgraph improves upon the baseline model (without using any subgraph) by $0.96$ absolute mAP, while adding only \emph{object-centric} subgraph gives a $0.64$ absolute mAP.
More importantly, our results show that the performance gain of each subgraph is complementary to each other.
To further validate the effectiveness of the \emph{object-centric} subgraph, we show in \tabref{ablation}(d) the breakdown of \tabref{ablation}(c) in terms of the number of persons in the scene. The \emph{object-centric} subgraph is less effective for cases with few people. For example, if there is only one person, the \emph{object-centric} subgraph has no effect. For images with a moderate amount of persons (4-6), however, our \emph{object-centric} subgraph shows a clear $0.98$ mAP gain. As the number of persons getting larger (7+), the \emph{object-centric} subgraph shows a relatively smaller improvement due to clutter. Among the 2,867 testing images, 68\% of them have only 1-3 persons. As a result, we do not see significant overall improvement.

\para{Spatial-semantic representation.}
To demonstrate the advantage and effectiveness of the use of the abstract spatial-semantic representation, we show in \tabref{ablation}(b) the comparison with alternative features, e.g., word2vec (as used in \cite{Lu-ECCV-Prior}) or appearance-based features.
By using our spatial-semantic representation in the dual relation graph, we achieve $51.37$ mAP. This shows a clear margin over the other alternative options, highlighting the contribution of spatial-semantic representation.



\begin{figure*}[!t]
\centering
\includegraphics[width=\ftqa]{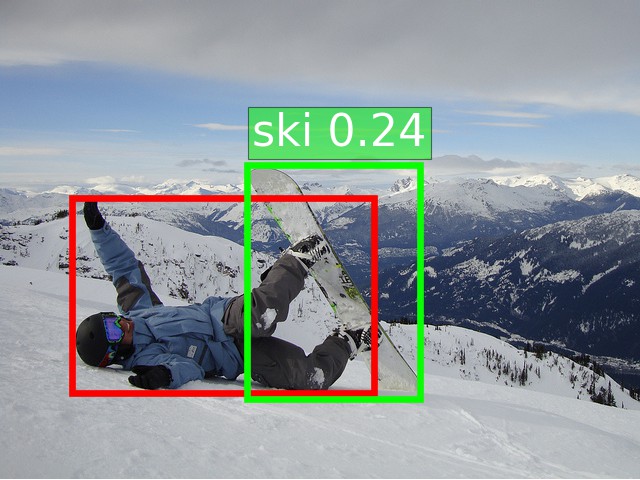}\hfill%
\includegraphics[width=\ftqa]{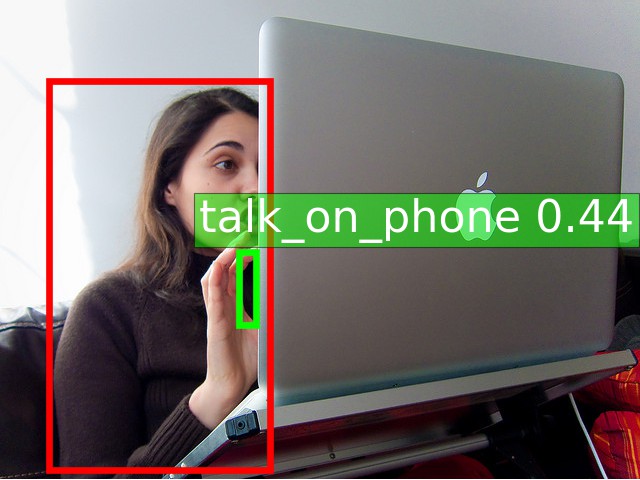}\hfill%
\includegraphics[width=\ftqa]{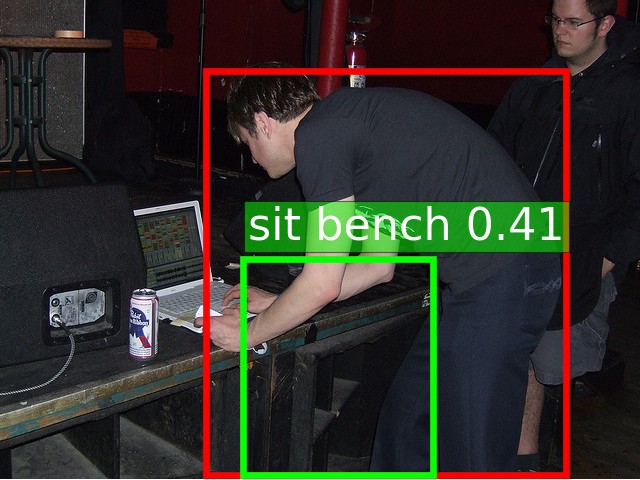}\hfill%
\includegraphics[width=\ftqa]{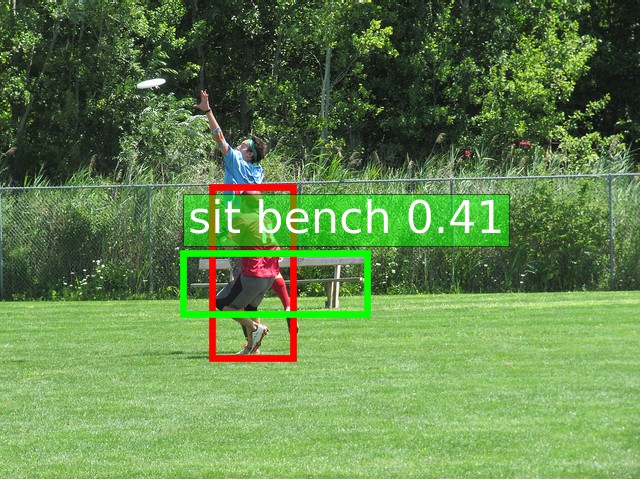}\hfill%
\includegraphics[width=\ftqa]{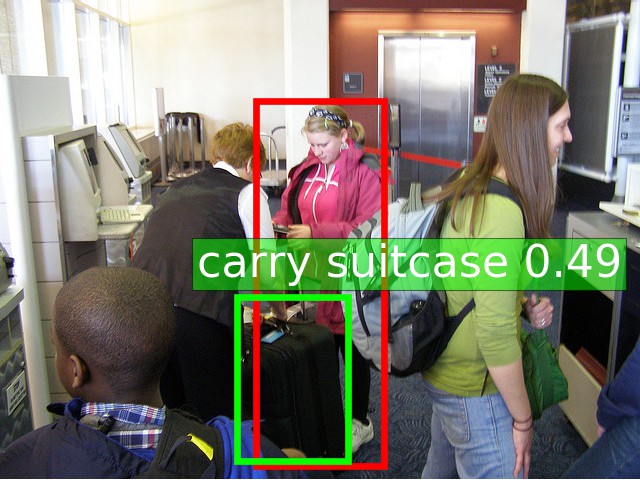}\hfill%
\includegraphics[width=\ftqa]{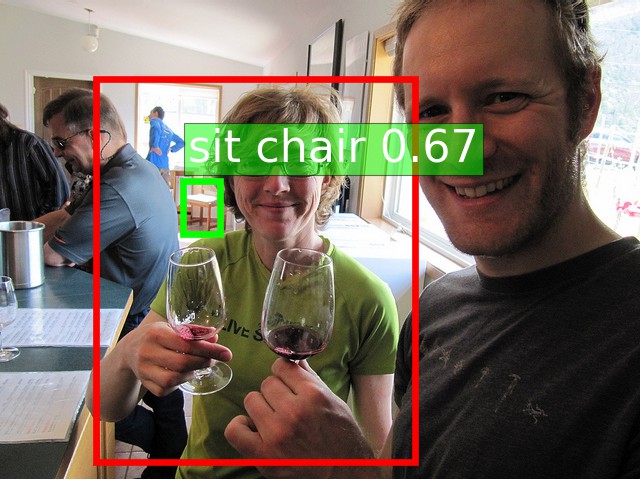}%
\mpage{0.45}{Incorrect object} \hfill
\mpage{0.45}{Incorrect action}
\caption{
\textbf{Failure cases of our method.}
}
\label{fig:failure}
\end{figure*}

\subsection{Limitations}
While we demonstrated improved performance, our model is far from perfect. 
Below, we discuss two main limitations of our approach, with examples in \figref{failure}.
%

First, we leverage the off-the-shelf object detector to detect object instances in an image. 
The object detection does \emph{not} benefit from the rich contextual cues captured by our method.
%
We believe that a joint end-to-end training approach may help reduce this type of errors. 

%
%
Second, our model may be confused by plausible spatial configuration and predicts incorrect action. 
In the third image, our model predicts that the person is sitting on a bench even though our model confidently predicts this person is standing and catching a Frisbee.
Capturing the statistics of co-occurring actions may resolve such mistakes.

\section{Conclusions}
\label{sec:conclusions}
In this paper, we present a Dual Relation Graph network for HOI detection. 
Our core idea is to exploit the \emph{global object layout} as contextual cues and use a \emph{human-centric} as well as an \emph{object-centric} subgraph to propagate and integrate rich relations among individual HOIs.
We validate the efficacy of our approach on two large-scale HOI benchmark datasets and show our model achieves a sizable performance boost over the state-of-the-art algorithms.
We also find that using the abstract spatial-semantic representation alone (i.e., without the appearance features extracted from a deep CNN) yields competitive accuracy, demonstrating a promising path of activity understanding through visual abstraction.

\para{Acknowledgements} We thank the support from Google Faculty Award.

\clearpage
%
%
\bibliographystyle{splncs04}
\bibliography{main}
\end{document}